\documentclass[a4paper,fleqn]{cas-dc}

\usepackage[numbers]{natbib}
\usepackage{subfigure}
\usepackage{float}
\usepackage{amsmath}
\usepackage[ruled]{algorithm2e}
\usepackage{soul}
\usepackage{bm}
\usepackage{color, xcolor}
\soulregister\cite7
\soulregister\citep7
\soulregister\ref7
 
\usepackage{tabularx}
\usepackage{graphicx}
\usepackage{amssymb}
\usepackage{amsthm}
\usepackage{enumitem}

\theoremstyle{plain}
\newtheorem{theorem}{Theorem}[section]
\newtheorem{proposition}[theorem]{Proposition}

\theoremstyle{definition}
\newtheorem{definition}[theorem]{Definition}

\theoremstyle{remark}

\def\tsc#1{\csdef{#1}{\textsc{\lowercase{#1}}\xspace}}
\tsc{WGM}
\tsc{QE}
\newcommand{\A}{\mathbf{A}}
\newcommand{\B}{\mathbf{B}}

\begin{document}
\let\WriteBookmarks\relax
\def\floatpagepagefraction{1}
\def\textpagefraction{.001}

\shorttitle{}
\shortauthors{}
\title [mode = title]{Robust Multimodal Sentiment Analysis via Double Information Bottleneck}



%






\author[1,2]{\textcolor{black}{Huiting Huang}}\ead{huiting.huang@stu.xjtu.edu.cn}
\author[1,2]{\textcolor{black}{Tieliang Gong}}  \cormark[1] \ead{adidasgtl@gmail.com}
\author[3]{\textcolor{black}{Kai He}}\ead{kai_he@nus.edu.sg}
\author[5]{\textcolor{black}{Jialun Wu}} \ead{jialunwu@nwpu.edu.cn}
\author[4]{\textcolor{black}{Erik Cambria}}\ead{cambria@ntu.edu.sg}
\author[3]{\textcolor{black}{Mengling Feng}}\ead{ephfm@nus.edu.sg}

\address[1]{School of Computer Science and Technology, Xi’an Jiaotong University, 710049, China}
\address[2]{Shaanxi Provincial Key Laboratory of Big Data Knowledge Engineering, Xi’an Jiaotong University, Xi’an, Shaanxi, 710049, China}
\address[3]{Saw Swee Hock School of Public Health, National University of Singapore, 119077, Singapore}
\address[4]{College of Computing and Data Science, Nanyang Technological University, 639798, Singapore}
\address[5]{School of Computer Science, Northwestern Polytechnical University, 710049, China}
  
\cortext[1]{Corresponding author.}



\begin{abstract}
Multimodal sentiment analysis has received significant attention across diverse research domains. 
Despite advancements in algorithm design, existing approaches suffer from two critical limitations: insufficient learning of noise-contaminated unimodal data, leading to corrupted cross-modal interactions, and inadequate fusion of multimodal representations, resulting in discarding discriminative unimodal information while retaining multimodal redundant information. To address these challenges, this paper proposes a Double Information Bottleneck (DIB) strategy to obtain a powerful, unified compact multimodal representation. Implemented within the framework of low-rank Renyi’s entropy functional, DIB offers enhanced robustness against diverse noise sources and computational tractability for high-dimensional data, as compared to the conventional Shannon entropy-based methods. The DIB comprises two key modules: 1) learning a sufficient and compressed representation of individual unimodal data by maximizing the task-relevant information and discarding the superfluous information, and 2) ensuring the discriminative of multimodal representation through a novel attention bottleneck fusion mechanism. Consequently, DIB yields a multimodal representation that effectively filters out noisy information from unimodal data while capturing inter-modal complementarity. Extensive experiments on CMU-MOSI with 2.2K samples, large-scale CMU-MOSEI with 22.9K, CH-SIMS with 2.3K, and MVSA-Single with 4.5K validate the effectiveness of our method. The model achieves 47.4\% accuracy under the Acc-7 metric on CMU-MOSI and 81.63\% F1-score on CH-SIMS, outperforming the second-best baseline by 1.19\%. Under noise, it shows only 0.36\% and 0.29\% performance degradation on CMU-MOSI and CMU-MOSEI respectively. The findings also uncover valuable potential for future work in addressing the challenges of global supervision constraints and reasoning over abstract visual semantics. The code is released on \url{https://github.com/Taylor-HHT/DIB}.



\end{abstract}


\maketitle





\section{Introduction}
Humans inherently experience the world in a multimodal manner, employing all five senses (i.e. sight, sound, smell, touch, and taste) to gather and interpret information for nuanced understanding and responses. Even when sensory signals are unreliable, humans excel at deriving meaningful insights from imperfect multimodal inputs, effectively reconstructing the context of events~\cite{camben,rideaux2021multisensory,mao2022biases,he2022meta,blending}. Advances in sensory technology now replicate this ability, enabling the seamless collection of diverse data streams for computational analysis. It paves the way for in-depth downstream applications such as multimodal sentiment analysis (MSA). MSA, which analyzes and interprets human sentiments across diverse modalities, has gained prominence in a range of applications, including
user engagement~\citep{somarathna2023exploring,yu2024artificial}, personalized recommendations~\citep{wu2024promise,he2021construction}, conversational systems~\cite{malandri2023convxai,liukno,lialea} and risk assessment~\cite{zambrano2023opportunistic,he2025survey,wu2023megacare}.
\begin{figure}
  \centering
  \includegraphics[width=\linewidth]{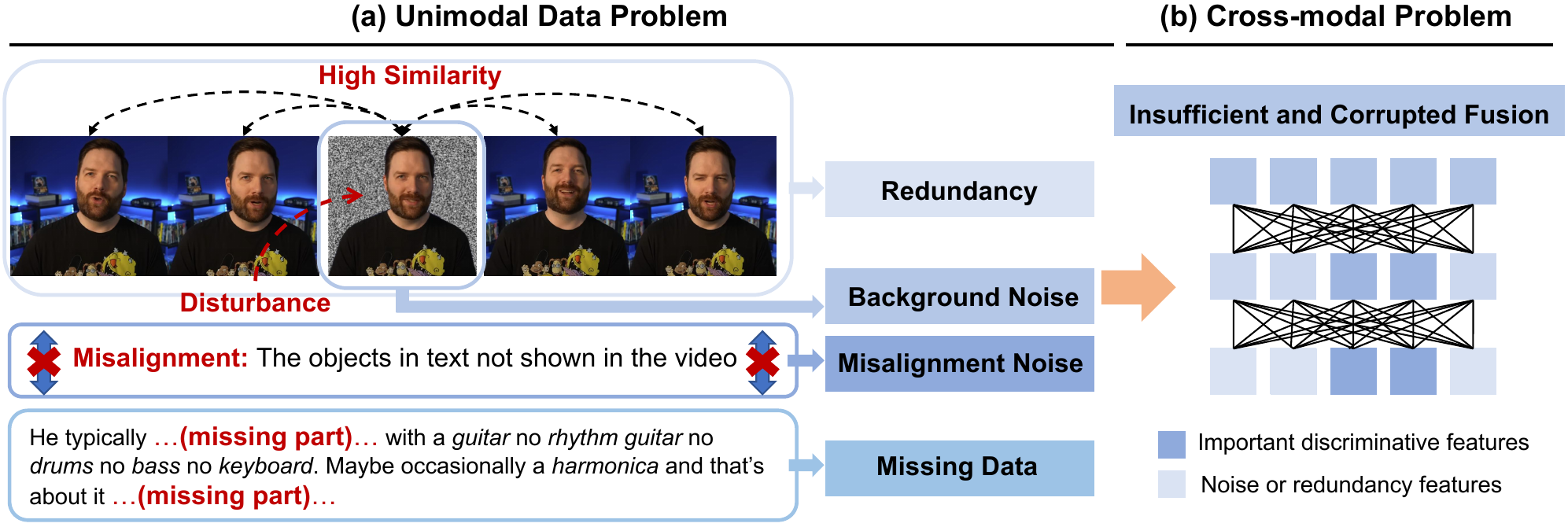}
  \caption{A visual-text pair example illustrating the unimodal contamination and cross-modal fusion problems: a) contaminated unimodal data includes redundancy (e.g. high similarity between consecutive frames), background noise, modality misalignment (e.g. objects mentioned in the transcript are not visible in the video) and missing data. b) the above contaminated unimodal data leads to corrupted and insufficient cross-modal interaction.}
  \label{fig:noise demonstration}
\end{figure}

\begin{figure*}[!ht]
\centering
\includegraphics[scale=0.45]{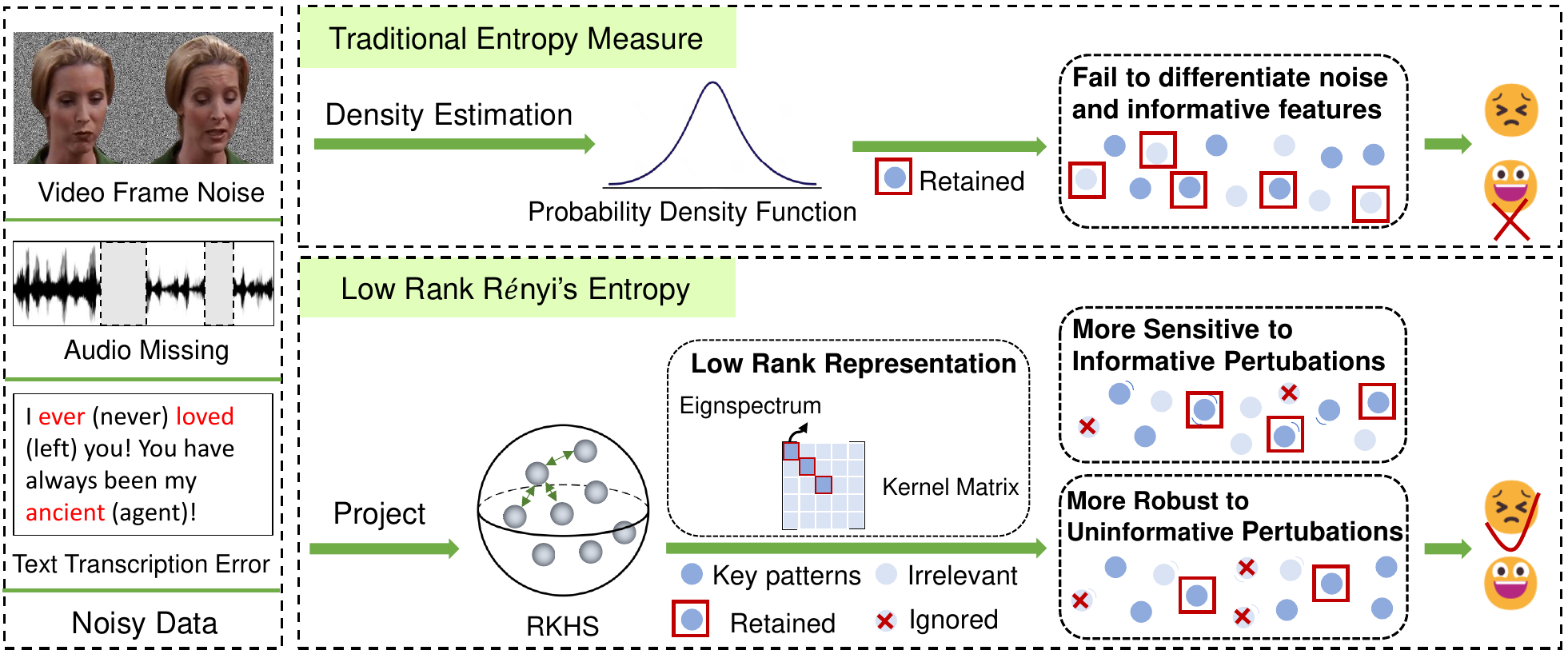}
\caption{Comparison of traditional entropy measure and low-rank R\'enyi’s entropy. Darker colors represent key patterns of features, while lighter colors indicate irrelevant features. The low-rank constraint in the R\'enyi's entropy ensures that only a few principal patterns in the multimodal features are retained in the representation, capturing the most salient features while ignoring the irrelevant and noisy parts.}
\label{fig:low-rank entropy}
\end{figure*}
Significant efforts have focused on extracting and integrating semantic information from different modalities to identify sentiments expressed in multimodal data~\cite{kim2023aobert,maauec,wancim}. In contrast to a single-modality sentiment analysis, cross-modal data inherently presents diverse and heterogeneous information, offering extra cues for emotional disambiguation in the context of sentiment analysis. 

For instance, lexical ambiguity in text, where words may convey multiple meanings and emotional connotations depending on context, underscores the importance of additional modalities~\cite{zadeh2017tensor,lin2024has}. In such cases, supplementary modalities provide clarifying contextual cues that aid in disambiguating the intended emotion. More complex scenarios, such as sarcasm~\cite{castro2019towards}, further indicate that unimodal analysis is inadequate in accurately capturing the underlying sentiment. Therefore, multimodal approaches, which integrate cues from various modalities, become essential.
However, the primary challenge in multimodal learning lies in effective integration of salient information across diverse modalities. 

Researchers have proposed various strategies to tackle the challenge of cohesive multimodal representation learning. Xu et al.~\cite{xu2018attngan} introduced an unsupervised approach that aligns modality-specific embeddings within a shared representation space. Their method employs an attention mechanism to extract essential features, capitalizing on the inherent similarities between image and text modalities. Conversely, Nguyen et al.~\cite{nguyen2019multi} proposed a supervised multi-task framework. By dynamically alternating between tasks and adaptively updating parameters, their approach captures both shared and modality-specific features, enabling the effective learning of a unified multimodal representation.

Albeit substantial progress has been made with the methods discussed above, existing approaches still suffer from two critical limitations, as depicted in Figure~\ref{fig:noise demonstration}: 1) insufficient learning of noise-contaminated unimodal data (e.g. redundancy, background interference, modal inconsistency, missing data), leading to corrupted cross-modal interactions, and 2) inadequate fusion of multimodal representations, resulting in discarding discriminative unimodal information while retaining multimodal redundant information. As demonstrated by~\cite{han2021improving,zhang2023learning}, sentiments extraction from multimodal data can be effectively accomplished by focusing on the most salient features rather than processing the entire feature set, as the inclusion of irrelevant or redundant features would degrade model performance.

To address these issues, we seek to learn robust unimodal representations that retain critical information while suppressing redundancy, and to design a compact yet expressive fusion mechanism for effective multimodal integration. To this end, we propose a Double Information Bottleneck (DIB) framework and validate it across diverse datasets. The DIB framework demonstrates adaptability to informative distribution shifts while maintaining resilience against uninformative perturbations, including measurement errors and background noise. At the core of our approach lies the 
low-rank R\'enyi's entropy functional~\citep{dong2023robust}, which extends traditional Shannon entropy while offering direct computability from empirical data without requiring knowledge of the underlying probability density function (PDF). 

As shown in Figure~\ref{fig:low-rank entropy}, traditional entropy measures (e.g. Shannon entropy, R\'enyi's entropy) rely on accurate density estimation of the underlying PDF, which is often challenging in high-dimensional settings. Moreover, these measures treat all features equally, failing to differentiate informative patterns from noise \cite{yu2019multivariate}, resulting in limited robustness. In contrast, low-rank R\'enyi's entropy avoids explicit density estimation by operating directly on data samples projected into a reproducing kernel Hilbert space (RKHS), where sample similarities are captured by a normalized kernel Gram matrix, whose eigenspectrum approximates the data distribution. The low-rank R\'enyi's entropy employs low-rank approximation by retaining only the top-k largest eigenvalues of the Gram matrix, intrinsically suppressing irrelevant or noisy components while preserving the principal patterns, thereby significantly improving robustness and performance when employed in the Low-rank Rényi's entropy-based information bottleneck (LRIB) in place of the original information bottleneck (IB).

Moreover, recent studies have revealed that conventional multimodal fusion methods often underperform when processing real-world, low-quality multimodal data, particularly in the presence of noise~\cite{xu2022different} or corruption~\cite{sun2023efficient} multimodal inputs. Drawing inspiration from~\citep{nagrani2021attention}, we design an attention bottleneck fusion module that mitigates cross-modal redundancy and irrelevance. This bottleneck architecture constrains information flow through low-capacity embeddings, effectively filtering out redundant and noisy information while preserving essential cross-modal patterns.

Generally, the proposed DIB framework comprises two primary components:

\begin{enumerate}
\item Unimodal Learning Module: generating intra-modal representations from unimodal inputs using LRIB;

\item Multimodal Learning Module: constructing a unified, compact inter-modal representation through our novel attention bottleneck fusion mechanism, followed by processing with LRIB.
\end{enumerate}

Our research makes three principal contributions:
\begin{itemize}

\item We develop the DIB framework incorporating low-rank Rényi entropy functional, offering enhanced robustness over conventional Shannon entropy while maintaining computational tractability in high-\\dimensional settings.


\item Our framework enables unified compact representation learning through joint optimization of unimodal feature compression and cross-modal correlation preservation, effectively capturing essential information while eliminating redundant components.

\item Through comprehensive experimentation on benchmark MSA datasets, we demonstrate DIB's superior performance compared to state-of-the-art methods, particularly highlighting its robust performance across varying noise conditions.
\end{itemize}

\section{Related Work}

\subsection{Multimodal Sentiment Analysis}
MSA has garnered significant attention due to its ability to leverage complementary information from various modalities such as visual and acoustic cues, as well as text~\cite{pandey2023progress}. Recent extensive work has focused on designing various fusion strategies to extract complementary information between different modalities for interpreting the latent sentiment~\cite{tsai2019multimodal,yu2022hierarchical,kim2023aobert}. Multimodal fusion techniques in sentiment analysis are typically categorized into feature-level, decision-level, and hybrid fusion. The main differences between them lie in the stage at which the modalities are combined: feature-level fusion integrates the feature information across multiple modalities at the early input level~\citep{park2016multimodal}, while decision-level fusion merges the prediction of individual modality at a later stage~\citep{cai2015convolutional}, and hybrid fusion integrates both feature-level and decision-level strategies to balance the strengths of both approaches~\cite{poria2016fusing}. However, these methods still struggle to effectively capture complex inter-modal relationships, particularly in the presence of noise.

Nowadays, attention-based fusion methods have attracted considerable interest, which models dynamic and complex interaction between modality-specific representations by leveraging attention weights~\cite{tsai2019multimodal,cheng2021multimodal}. Tsai et al. propose MulT~\citep{tsai2019multimodal}, a crossmodal transformer that relies solely on attention to handle unaligned multimodal sequences by focusing on relevant signals without requiring explicit alignment. AOBERT~\cite{kim2023aobert} captures the essential dependencies and relationships between modalities by simultaneously learning to mask and align multimodal data during BERT pre-training. HIMT~\cite{yu2022hierarchical} leverages hierarchical attention mechanisms to first model the aspect-text and aspect-image interactions, followed by capturing the text-image interactions. ALMT~\cite{zhang2023learning} designs a language-dominant learning module, which dynamically updates the text representation by calculating and integrating attention-weighted audio and visual features. However, while designing complex fusion strategies to obtain excellent results, the aforementioned multimodal fusion techniques often lead to potential redundancy and noise information retained in the learned high-dimensional representation, with large-scale attention calculations further exacerbating the risk of retaining irrelevant information. To address these issues, we introduce an advanced attention bottleneck fusion mechanism to enhance the effectiveness of multimodal integration by constraining the information flow.

\subsection{Information Bottleneck in Deep Learning}
To reduce redundancy and noise information in latent features, a series of deep learning methods driven by IB have been proposed, facilitating the learning of sufficient and compressed representation~\citep{han2021improving,hoang2022multimodal,hu2024survey}. The IB theory for deep learning is first presented in~\cite{tishby2015deep}, with supporting empirical study~\cite{shwartz2017opening}. VIB~\cite{alemi2016deep} provides a gradient-based optimization method to solve the IB Lagrangian in deep neural network. Since then, there are numerous works which adopt this concept as a design tool, including classification~\cite{achille2018information} and generative models~\cite{higgins2017beta}. Amjad et al.~\cite{amjad2019learning} design regularizers on latent representation to alleviate the optimization problem and the invariance of IB functional under bijections problem. Wan et al.~\cite{wan2021multi} employ IB theory and mutual information (MI) implemented by variational reference on unsupervised multi-view representation learning problem, thereby learning intra-view intrinsic information and inter-view shared structure.

Several IB-inspired approaches in MSA have emerged. MMIM~\cite{han2021improving} hierarchically maximizes MI between the fusion representation and unimodal inputs to retain task-relevant information, but it lacks explicit mechanisms to suppress modality-specific noise, especially in low-quality inputs. CMIMH~\cite{hoang2022multimodal} leverages MI to learn binary hash codes for efficient cross-modal retrieval. However, its discrete representation may limit expressive capacity for sentiment modeling and lacks explicit mechanisms to handle noisy or redundant inputs. MIB~\cite{mai2022multimodal} introduces three IB-based fusion strategies and optimizes mutual information between labels and learned representations, yet it heavily relies on variational estimation and Shannon entropy, which can be computationally expensive and less robust to high-dimensional noise. Unlike the aforementioned methods, our method leverages low-rank R\'enyi’s entropy, which generalizes Shannon entropy and offers two distinct advantages. First, it avoids explicit density estimation by operating on kernel-based sample similarities, making it more tractable in high-dimensional spaces. Second, the low-rank approximation inherently filters out irrelevant or noisy components by retaining only dominant eigenvalues. This theoretical foundation allows our method to suppress spurious patterns while preserving the most salient structures.

%


\section{Preliminaries}
Throughout this work, we denote random variables by capitalized letters $(X)$ and their specific realizations by lowercase letters $(x)$. Let $P_X$ be the distribution of a random variable $X$ and $P_{X|Y}$ be the conditional distribution of $X$ conditioned on $Y$. Let $H(X)$ be Shannon's entropy, and $I(X;Y)$ be the mutual information between random variables $X$ and $Y$. $\mathrm{KL}(P \| Q)$ denotes the Kullback–Leibler divergence of $P$ with respect to $Q$. 

\subsection{Entropy Measures}
We introduce the theoretical foundation of R\'enyi's entropy~\cite{renyi1961measures} and its matrix-based formulation. Our approach builds on a low-rank approximation of matrix-based R\'enyi's entropy to quantify information in a robust and data-driven manner. For a random variable $X$ with density $p(x)$, the $\alpha$-order R\'enyi's entropy is defined as:
\begin{equation}
H_{\alpha}(X) = \frac{1}{1 - \alpha} \log_2 \int p(x)^\alpha dx,\,\, \alpha > 0, \alpha \neq 1.
\end{equation}
As $\alpha \rightarrow 1$, it converges to Shannon entropy. However, it is easy to see that both Shannon and classical R\'enyi's entropy require accurate estimation of the underlying probability density function (PDF), which becomes impractical in high-dimensional, sample-driven settings due to the curse of dimensionality. To overcome this, a matrix-based formulation~\cite{sanchez2014measures} defines R\'enyi entropy directly from samples via kernel matrices. Given samples $\{x_i\}_{i=1}^n$ and a positive definite kernel $\kappa$, one constructs the normalized Gram matrix \textbf{$A$}, with entries:
\begin{equation}
\mathbf{A}_{ij} = \frac{1}{n} \cdot \frac{K_{ij}}{\sqrt{K_{ii}K_{jj}}}, \,\, K_{ij} = \kappa(x_i, x_j),
\end{equation}
where \textbf{$A$} is positive semi-definite with $\text{tr}(A) = 1$. The matrix-based R\'enyi's entropy is then:
\begin{equation}
H_{\alpha}(\mathbf{A}) = \frac{1}{1 - \alpha} \log_2 \left( \textstyle\sum_{i=1}^n \lambda_i^\alpha(\mathbf{A}) \right),
\end{equation}
where $\lambda_i(\mathbf{A})$ denotes the $i$-th eigenvalue of $\mathbf{A }$. This formulation avoids explicit density estimation, and naturally extends to mutual information and conditional entropy~\cite{yu2019multivariate}.
In this work, we adopt a low-rank approximation of $H_{\alpha}(A)$ via truncated eigenspectrum, enabling robust information measurement in noisy and high-dimensional settings.

\begin{figure*}[h]
\centering
\includegraphics[scale=0.5]{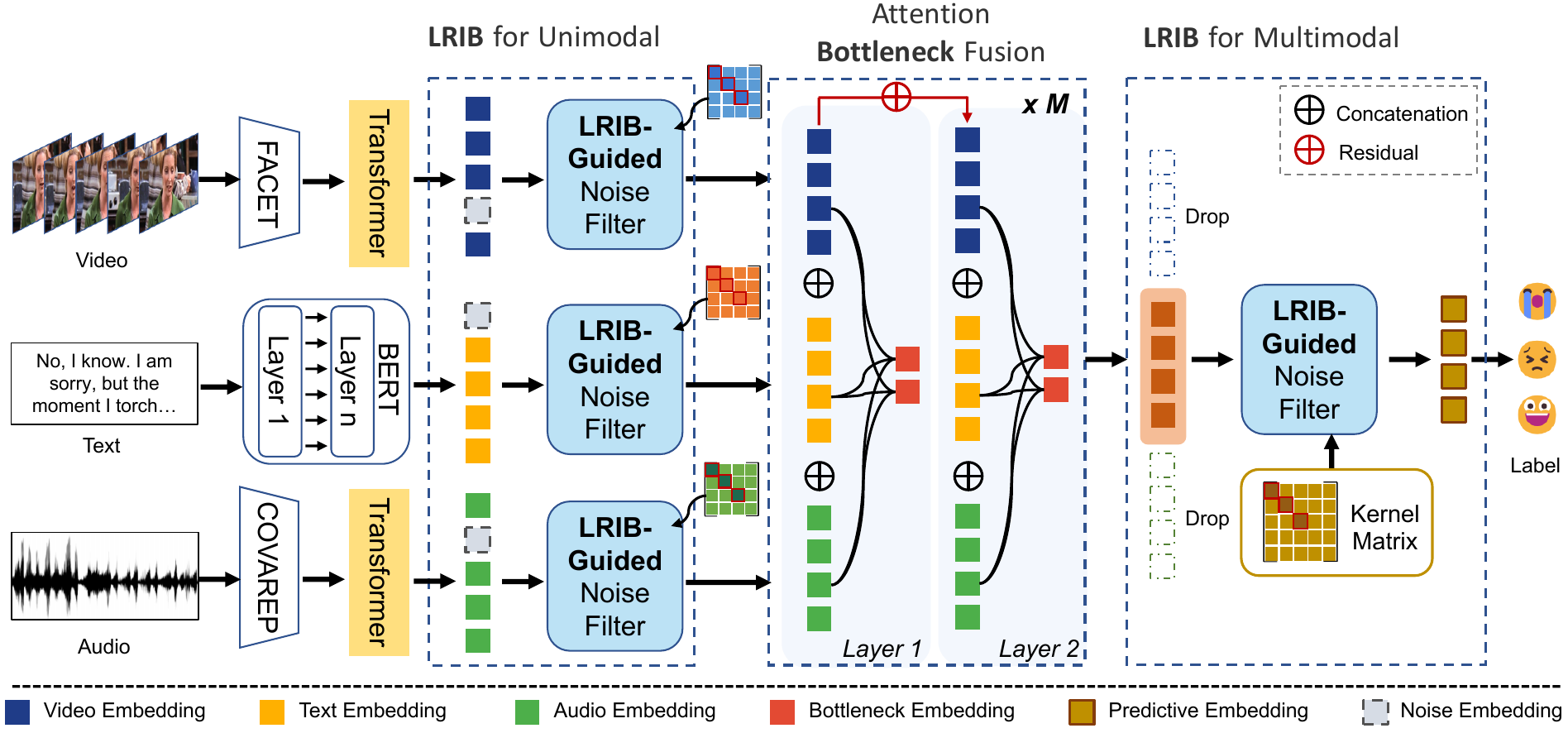}
\caption{The architecture of the proposed DIB model. After feature extraction, LRIB-guided representation learning modules act as the noise filter at both unimodal and multimodal levels. In addition, attention bottleneck fusion sifts information to produce a unified and compact representation.}
\label{fig:architecture}
\end{figure*}


\subsection{Information Bottleneck}
  Rooted in rate-distortion theory, the IB was proposed in~\cite{tishby2000information} as a principled approach for extracting relevant information from an observed signal about a target one. For a pair of correlated random variables $(X, Y)$, IB has emerged as an information-theoretic framework in numerous practical deep learning realms including computer vision~\cite{peng2018variational}, reinforcement learning~\cite{goyal2019infobot}, and natural language processing~\cite{han2021improving}. 
  
  Given a data space $X$ with a fixed probability measure $p(x)$, the IB method seeks to learn an effective quantized representation $T$ that satisfies two key objectives: (1) Compression: The representation $T$ should discard irrelevant details and noise from $X$, which is measured by
\begin{equation}
I(X ; T)=\int p(x, t) \log \frac{p(t | x)}{p(t)} \mathrm{d} x \mathrm{~d} t,
\end{equation}
where a smaller $I(X;T)$ indicates a higher degree of compression. (2) Relevance Preservation: While compressing $X$, $T$ should retain as much relevant information as possible about a target variable $Y$, ensuring its effectiveness for downstream tasks. This is quantified by 
\begin{equation}
  \begin{aligned}
    I(T;Y) & =\int p(y, t) \log \frac{p(y, t)}{p(y) p(t)} \mathrm{d} y \mathrm{~d} t \\
    & =\int p(y, t) \log \frac{p(y | t)}{p(y)} \mathrm{d} y \mathrm{~d} t \leq I(X ; Y),
  \end{aligned}
  \label{eq:IZY}
\end{equation}
where the inequality follows from the Data Processing Inequality \cite{thomas2006elements}. The IB objective aims to maximize the preserved information about $Y$ while minimizing the amount of information about $X$, thus imposing a compression constraint, given by
\begin{equation}
\mathcal{L}_{IB}=I(X;T)-\beta I(T;Y),
\label{eq:IB principal}
\end{equation}
where $\beta \in [0, \infty)$ serves as a Lagrange multiplier controlling the trade-off between compression and information preservation. By adjusting $\beta$, researchers can explore various operating points along the compression-relevance curve.

\subsection{Modality Feature Extraction}
\label{sec:Modality Feature Extraction}
We illustrate the process of transforming raw multimodal data into embeddings that are used in our proposed model.

\textbf{Text Modality.} Each sample of the multimodal input consists of a single utterance from a speaker. In order to obtain rich semantic and contextual information across a sequence of raw words, several useful techniques are available, particularly transformer-based pre-trained language models. For a fair comparison, we follow previous works~\cite{mai2022multimodal,lin2023ps}, which employ BERT \cite{devlin2018bert} from the open-source Transformers library to preprocess and extract word-level features. Specifically, we use \textit{bert-base-uncased} model for CMU-MOSI and CMU-MOSEI dataset and \textit{bert-base-chinese} model for CH-SIMS dataset. For MVSA-Single dataset, we use the ViT-B/16 model to extract textual features.

\textbf{Visual Modality.} For video datasets, we encode visual data from multiple video frames utilizing the FACET~\cite{imotions2017facet} tool which extracts facial expression features including facial action units, facial landmarks, and head pose. A dedicated Transformer is then employed to generate a visual embedding $E^v$. In particular, we use the \textit{OpenFace 2.0} toolkit~\cite{openface2018} to extract a variety of features including 17 facial action units, 68 facial landmarks, and several head and eye-related metrics. For MVSA-Single image-text dataset, we use the ViT-B/16 model to extract image features.

\textbf{Audio Modality.} In our paper, the audio feature embeddings $E^a$ are primarily extracted using the COVAREP~\cite{degottex2014covarep} tool, which offers a spectrum of acoustic features, including fundamental frequency, normalized amplitude quotient, 12 Mel-frequency cepstral coefficients (MFCCs), followed by the dedicated audio Transformer. Specifically, we utilize the \textit{Librosa} Python package~\cite{mcfee2015librosa} to extract key features, such as the logarithmic fundamental frequency, 12 Constant-Q chroma features, and 20 MFCCS.

\section{Method}
\label{sec:method}
In this section, we first define the task, followed by the definition and properties of our proposed LRIB. Subsequently, we will introduce the architecture of DIB, as illustrated in Figure~\ref{fig:architecture}, which comprises three modules: 1) to extract discriminative representations of each modality, we propose the LRIB-based unimodal learning method; 2) to avoid the interaction of noise and redundant information across different modalities, we design a novel attention bottleneck fusion mechanism with bottleneck embeddings; 3) to further enhance the effectiveness of the fused representation, we apply LRIB to the multimodal representation learning and employ the enriched representation of textual modality for target learning tasks.

\subsection{Task Definition}
The downstream task in our work is multimodal sentiment analysis, and the key point lies in automatically and effectively integrating diverse modalities, such as text (t), acoustic (a), visual (v), to identify the underlying sentiment of a given multimodal sample. Let the multimodal input be denoted as $X = {X^m}_{m \in \mathcal{M}}$, where $\mathcal{M}$ is the set of available modalities (e.g., ${t, a, v}$), and each $X^m \in \mathbb{R}^{l_m \times d_m}$ represents the feature sequence from modality $m$, with $l_m$ indicating sequence length and $d_m$ the feature dimension. We then obtain modality-specific embeddings $E^m \in \mathbb{R}^{l_m \times d_m}$ accordingly. The whole feature extraction process to gain the embedding $E^m$ from raw multimodal inputs is as described in Section~\ref{sec:Modality Feature Extraction}. Each modality trains its own parameters of the encoding module, rather than merging them early in the process, thereby preserving the distinct information of each modality.

The distinct modality information would then be fused into a comprehensive and unified representation $T \in \mathbb{R}^{d}$, which plays a crucial role in classifying each sample into a sentiment label $y_i$ and finally outputting a label sequence $Y=\{y_1,y_2,...,y_{|C|}\}$, where $C$ denotes the predefined set of categories. The prediction for discrete sentiment categories could be formulated as:
\begin{equation}
  \hat{y}=\operatorname{argmax}_{y_i\in\mathcal{C}}\operatorname{P}_\Phi(y_i | T),
\end{equation}
where $\Phi$ represents the model parameters. 

We also estimate the sentiment intensity score $\hat{y}\in \mathbb{R}$, represented by
\begin{equation}
  \hat{y}=\operatorname{P}_\Phi(T).
\end{equation}
By addressing this problem, we develop robust methodologies that enhance the understanding of complex emotions conveyed through multimodal data. The main notations used in section~\ref{sec:method} are summarized in Table~\ref{tab:symbols} for the convenience of reference.

\begin{table*}[h]
\caption{Description of main symbols used in Section~\ref{sec:method}. }
\centering
\begin{tabular}{c|l}
\hline
\textbf{Symbol} & \textbf{Description} \\
\hline
$X^m$ & Multimodal data input \\
$E^m$ & Embeddings of each modality \\
$Z^m$ & The encoded unimodal representations\\
$Z^m_M$ & The encoded unimodal representations after $M$ layers fusion process\\
$Z$ & The multimodal representation obtained by applying ReLU to the fused text features\\
$\Tilde{Z}$ & The final multimodal representation learned through LRIB\\
$Y^m$ & The true label of unimodal samples, which we substitute with the overall multimodal label $Y$\\
$\hat{Y}^m,\hat{Y}$ & The predicted label of unimodal and multimodal representation\\
$\beta_m,\beta$ & The unimodal, multimodal Information Bottleneck Lagrange multiplier, respectively \\
$X_A,E_A,Z_A,\Tilde{Z}_A,Y_A$ & The normalized kernel matrix constructed from $\{X\}_{i=1}^n,\{E\}_{i=1}^n,\{Z\}_{i=1}^n,\{\Tilde{Z}\}_{i=1}^n,\{Y\}_{i=1}^n $\\
$\mu^m,\sigma^m$ & The mean, standard deviation for modeling Gaussian distribution in unimodal variational encoder\\
$\epsilon^m$ & The unimodal stochasticity introduced in reparameterization trick\\
$\mu,\sigma,\epsilon$ & The variational encoder parameters for multimodal representation\\
$\phi^m,\phi$ & Parameters of unimodal and multimodal variational encoders, respectively\\
$\Phi$ & Parameters of the whole model\\
\hline
\end{tabular}
\label{tab:symbols}
\end{table*}

\subsection{Low-rank R\'enyi’s entropy-based Information Bottleneck (LRIB)}
  We employ the low-rank R\'enyi’s entropy-based information bottleneck instead of original IB to obtain a compact and informative representation, built upon the principles of low-rank matrix-based R\'enyi’s entropy. Next, we will introduce low-rank R\'enyi's entropy and outline the definition and key properties to provide a deeper understanding of LRIB.

\begin{definition}[\textbf{Low Rank R\'enyi's Entropy}]
  We adopt the following low-rank formulation of matrix-based R\'enyi's entropy to efficiently estimate information from data via kernel eigenvalues. Let $\kappa:\mathcal{X}\times\mathcal{X}\mapsto\mathbb{R}$ be an infinitely divisible kernel~\cite{bhatia2006infinitely}, which maps pairs of elements from $\mathcal{X}$ to real numbers. Given $\{X_i\}_{i=1}^n\subset\mathcal{X} $ and an integer $k \in [1,n-1]$, the low-rank R\'enyi's $\alpha$-order entropy ($\alpha >0, \alpha \neq 1$) is defined by 
  \begin{equation}\label{LR_calculate}
  \begin{aligned}
    H_{\alpha}^{k}(\mathbf{A})=\frac{1}{1-\alpha}\log_{2}\biggl(\sum_{i=1}^{k}\lambda_{i}^{\alpha}(\mathbf{A})+(n-k)\lambda_{r}^{\alpha}(\mathbf{A})\biggr),
  \end{aligned}
  \end{equation}
  where $\A$ denotes the normalized kernel matrix constructed from $\{X_i\}_{i=1}^n$, $\lambda^\alpha_{i}(\A)$ denotes the $i$-th largest eigenvalue of $\A$, and $\lambda^\alpha_r(\A)=\frac1{n-k}\big(1-\sum_{i=1}^k\lambda_i^\alpha(\A)\big)$, representing the contribution of the remaining eigenvalues. 
\end{definition}  
  Note that the low-rank approximation of R\'enyi’s entropy takes advantage of the eigenvalues of the kernel matrix $\A$ to capture essential information. 
  The corresponding joint entropy, conditional entropy, and mutual information are defined by 
  \begin{eqnarray}
     H_{\alpha}^{k}(\A, \B) &=& H_{\alpha}^{k} \left( \frac{\A \circ \B}{tr(\A \circ \B)} \right), \label{eq:joint}\\
    H_{\alpha}^k (\A|\B) &=& H_{\alpha}^k (\A,\B) - H_{\alpha}^k (\B),\label{eq:conditional}\\ 
    I_{\alpha}^k (\A;\B) &=& H_{\alpha}^k (\A) + H_{\alpha}^k (\B) - H_{\alpha}^k (\A,\B),\label{eq:mutual}
  \end{eqnarray}
  where $\A \circ \B$ represents the Hadamard product of matrices $\A$ and $\B$, and $tr(\cdot)$ is the trace of the matrix. The entropy requires computing the Positive Semi-Definite (PSD) matrix $\A$, which has a time complexity of $\mathcal{O}(n^3)$ through eigenvalue decomposition algorithms. To mitigate computational burden, we employ Lanczos iteration techniques~\cite{lanczos1950iteration} to efficiently approximate the matrix, significantly reducing the time complexity to $\mathcal{O}(n^2s)$, where $s \ll n$ denotes the number of queried random vectors~\cite{dong2023robust}. 
  
  The Eq.~\ref{eq:mutual} outlines the method for calculating mutual information using low-rank R\'enyi's entropy measure, which will then be used in the IB equation. Note that the bold capital letters $\A, \B \in\mathbb{R}^{n\times n}$ in the above equations are both kernel square matrices. For convenience, we denote $X_A$, or expressions with subscript $A$ by the normalized kernel matrices constructed from the data $X$ or other variables.
  

To further clarify our theoretical motivation, we provide additional justification for adopting this entropy.
The equation~\ref{LR_calculate} can be viewed as a principled spectral approximation that retains the dominant eigenvalues of the normalized kernel matrix, thereby capturing the most informative subspace of the representation.
From the perspective of information theory, this design aligns with the definition of min-entropy, which measures information based on the most probable outcome, and with the Information Bottleneck principle, which seeks minimal sufficient statistics for prediction.
By focusing on the principal eigenspectrum, the low-rank form effectively suppresses noise-dominant components while preserving the intrinsic data structure encoded in high-energy directions.


\begin{definition}[\textbf{LRIB}]
  To extract compact and informative latent representations, we formulate an information bottleneck objective based on low-rank Rényi entropy, denoted as LRIB. Consider the observable input $X$ and the target $Y$, with $T$ representing the information related to $Y$ through $X$. Furthermore, $Y$ must not be independent of $X$, and $T$ is (a possibly randomized) function of X. Therefore they form the Markov chain $Y\leftrightarrow X\leftrightarrow T$. We define our LRIB (denoted simply as $\mathrm{IB}{^k_{\alpha}}(R)$ in the rest of the content) as followings:
  \begin{equation}
    \begin{aligned}
      \mathrm{IB}{^k_{\alpha}}(X,Y,k,R) &:={\max}\,\, I^k_\alpha(Y_A;T_A), \\
      \mathrm{s.t.} I^k_\alpha(X_A;T_A)&\leq R.
    \end{aligned}
  \label{eq:LRIB Definition}
  \end{equation}

  where $\alpha$ denotes the order of matrix-based R\'enyi's entropy, $k$ is the hyper-parameter of low-rank R\'enyi's entropy, and $R$ is maximum limit of information contained in $T$ of $X$. Specifically, this definition keeps a fixed number of bits from the original input $X$ (compression) while maximizing the amount of meaningful information about the relevant variable $Y$ (relevance). In other way, it provides a mechanism to predict the groundtruth precisely while accessing the minimal amount of information from the input to filter out the noise.
  By introducing a Lagrange multiplier $\beta$, the constrained optimization problem Eq.~\ref{eq:LRIB Definition} is equivalent to the following unconstrained one:
  \begin{equation}\label{eq:LRIB principal}
    \begin{aligned}
      \mathcal{L}_{LRIB}=I^k_\alpha(X_A;T_A)-\beta I(T;Y),
    \end{aligned}
  \end{equation}
\end{definition}

In practice, it is noteworthy that the first term is calculated by low-rank R\'enyi's entropy while the second term (i.e. $I(T;Y)$) is estimated using the variational method discussed in the subsequent section, as it represents the loss in the downstream task. Therefore, we denote them as distinct expressions $I^k_\alpha(\cdot)$ and $I(\cdot)$.

 By utilizing low-rank R\'enyi's entropy, we can 1) learn a more robust representation by preserving the most informative components in multimodal data while ignoring irrelevant and noisy parts by selecting the top $k$ largest eigenvalue of the calculated normalized Gram matrix, and 2) achieve computational efficiency for high-dimensional data while directly quantifying information measures without PDF, also retaining the properties of the conventional Shannon entropy employed by the original IB.

\begin{proposition}
  For any given $X$, $Y$, the mapping $\mathrm{IB}{^k_{\alpha}}$ have the following properties:
  \begin{enumerate}[label=(\alph*), ref=\theproposition(\alph*), leftmargin=0.75cm]
    \item $\mathrm{IB}{^k_{\alpha}}(R=0)=0.$
  \end{enumerate}
  \begin{proof}
    \renewcommand{\qedsymbol}{}
    In the optimization problem given by Eq.~\ref{eq:LRIB Definition}, $R=0$ implies that $X$ and $T$ are independent, as $I^k_\alpha(X_A;T_A)=0$, where mutual information cannot be negative. Furthermore, based on the Markov chain property, we can deduce that $I^k_\alpha(Y_A;T_A)=0$, which leads to the property (a).
    
  \end{proof}

  \begin{enumerate}[label=(\alph*), ref=\theproposition(\alph*), leftmargin=0.75cm]
    \setcounter{enumi}{1}
    \item $\mathrm{IB}{^k_{\alpha}}(R)= I^k_{\alpha}(X_A;Y_A),\quad \text{for any} \,\, R \geq H^k_{\alpha}(X_A).$
  \end{enumerate}
  \begin{proof}
    \renewcommand{\qedsymbol}{}
    Given that $I^k_{\alpha}(X_A;T_A) \leq H^k_{\alpha}(X_A) $ for any $X$ and $T$, the information constraint $I^k_{\alpha}(X_A;T_A) \leq R $ is automatically satisfied when $R \geq H^k_{\alpha}(X_A)$. Furthermore, a representation $T$ of $X$ is sufficient for $Y$ if and only if $I^k_\alpha(X_A;Y_A|T_A)=0 \Longleftrightarrow I^k_\alpha(X_A;Y_A)=I^k_\alpha(X_A;T_A)$. That is, $T$ and $Y$ contain identical information about $X$. According to the Data Processing Inequality (DPI)~\cite{thomas2006elements}, which holds for the IB framework, it follows that $I^k_{\alpha}(Y_A;T_A) \leq I^k_{\alpha}(X_A;T_A)$. This implies that $I^k_\alpha(X_A;Y_A)$ (and also $I^k_\alpha(X_A;T_A)$) serves as the upper bound for $I^k_\alpha(Y_A;T_A)$, thereby confirming that the choice of $I^k_\alpha(X_A;Y_A)$ is optimal.
  \end{proof}
  \begin{enumerate}[label=(\alph*), ref=\theproposition(\alph*), leftmargin=0.75cm]
    \setcounter{enumi}{2}
    \item $0\leq \mathrm{IB}^k_\alpha \leq \min \{R,I^k_\alpha(X_A;Y_A)\},\quad \text{for any} \,\, R\geq0.$
  \end{enumerate}
  \begin{proof}
    \renewcommand{\qedsymbol}{}
    For all $T$ satisfying Markov chain $Y \to X \to T$, which implies $T$ is a transformation of $X$ and the information that $T$ carries about $Y$ must flow through $X$, the upper bound on IB follows directly from the DPI. Specifically, $I^k_{\alpha}(Y_A;T_A)$ is constrained by both the information between $T$ and $X$, as well as the total information between $X$ and $Y$. This yields the upper bound: $I^k_{\alpha}(Y_A;T_A) \leq \min \{ I^k_{\alpha}(X_A;T_A), I^k_{\alpha}(X_A;Y_A) \}.$ By incorporating the constraint $I^k_\alpha(X_A;T_A) \leq R$, the DPI-derived bound leads to property (c).
  \end{proof}
  \begin{enumerate}[label=(\alph*), ref=\theproposition(\alph*), leftmargin=0.75cm]
    \setcounter{enumi}{3}
    \item $R \to \frac{\mathrm{IB}^k_\alpha(R)}{R}$ is non-increasing.
  \end{enumerate}  
  \begin{proof}
    \renewcommand{\qedsymbol}{}
    Define $M \subseteq \mathbb{R}^2$ as $M:=\{(I_\alpha^k(X_A;T_A), I_\alpha^k(Y_A;T_A)) \\| Y \to X \to T \}$. The empirical LRIB curve is obtained by training our DIB model, as illustrated in Figure~\ref{fig:IXZ_IYZ}. It can be observed that $M$ is convex. Consequently, the function $R \to \mathrm{IB}^k_\alpha(R)$ represents an upper bound for $M$, which implies that $\mathrm{IB}_\alpha^k(\cdot)$ is concave. 
    
    The monotonicity of the mapping $R \to \frac{\mathrm{IB}^k_\alpha(R)}{R}$ follows directly from the concavity of $\mathrm{IB}_\alpha^k(\cdot)$.
    
\begin{figure}
  \centering
  \includegraphics[width=\linewidth]{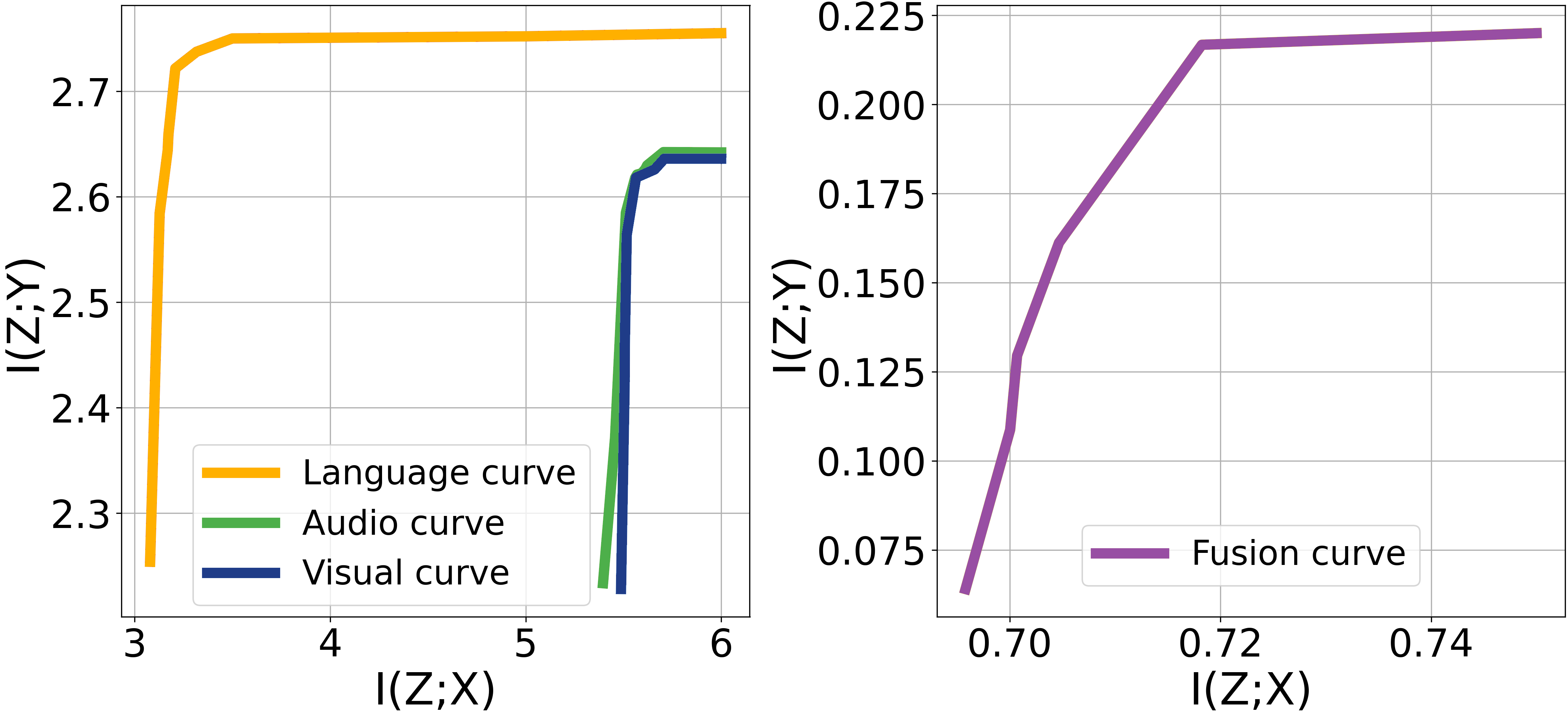}
  \caption{The empirical LRIB curve found by minimizing the LRIB Lagrangian of DIB model on CMU-MOSI dataset with varying $\beta$.}
  \label{fig:IXZ_IYZ}
\end{figure}    
   
  \end{proof}  
  \begin{enumerate}[label=(\alph*), ref=\theproposition(\alph*), leftmargin=0.75cm]
    \setcounter{enumi}{4}
    \item $\mathrm{IB}{^k_{\alpha}}(R) :={\sup} \,\,\, I^k_{\alpha}(Y_A;T_A), \quad \text{if} \,\,\, I^k_\alpha(X_A;T_A)= R.$
  \end{enumerate}  
  \begin{proof}
    \renewcommand{\qedsymbol}{}
    The strict monotony of $\mathrm{IB}_\alpha^k$ implies the optimization problem in Eq.~\ref{eq:LRIB Definition} reaches its solution when the inequality in the constraint becomes equality.
  \end{proof}
\end{proposition}


\subsection{LRIB for Unimodal Representation Learning}
\label{sec:Unimodal Representation Learning}
 For MSA task, effective representation learning is crucial to enhance the overall model performance~\cite{hazarika2020misa}. However, it faces challenges because emotion is inherently subjective and varies from person to person, which introduces a degree of uncertainty. More importantly, the multimodal input often contains noise and redundancy, further complicating efficient fusion. To alleviate the above problems, we employ variational encoder to obtain the stochastic Gaussian embedding of unimodal input and utilize LRIB to improve upon the traditional IB technique. 

Given the embedding $E^m$ of each modality and the true label $Y^m$, we optimize the following LRIB objective to extract the latent representation $Z^m$ for individual modality:
\begin{equation}\label{eq:LRIB' unidef}
\mathcal{L}_{Uni-LRIB}= \sum_{m\in \mathcal{M}}[I_{\alpha}^k(E^m_A;Z^m_A)-\beta_m I(Z^m;Y^m)].
\end{equation}
We use the overall label $Y$ as a substitute for $Y^m$ to guide learning in the unimodal context. This LRIB objective can obtain the minimal sufficient representation of unimodality by extracting the relevant information about the label $Y^m$ while eliminating the mutual information between $E^m$ and $Z^m$. Note that $Z^m$ is obtained after applying a separate variational encoder $\phi_m$ to each modality. Unlike traditional IB methods that require a fixed variational encoder, our design allows the encoder to be flexibly chosen based on the task. In this work, we adopt a variational encoder because it introduces stochasticity into $Z^m$, which helps the model capture individual variability and uncertainty in emotional expression. This probabilistic representation improves robustness and generalization in subjective tasks such as sentiment analysis.

To realize this variational formulation, we regard the unimodal encoders in our framework as probabilistic models that learn the distributions $p(Z^m|E^m)$ over the latent space. Specifically, we assume that $Z^m$ follows a Gaussian distribution conditioned on the input embedding $E^m$.
Each encoder is implemented as a multi-layer perceptron (MLP), which predicts the mean and log-variance of the Gaussian posterior. These parameters define the latent distribution from which $Z^m$ is sampled during training:
\begin{equation}
  \left[\mu^m, \log \sigma^m\right]=\operatorname{ReLU(}\operatorname{MLP}^m\left(E^m)\right).
\end{equation}
However,  the direct optimization of the  stochastic encoders $p(Z^m|E^m)$ is not trivial, as its gradients involve probabilistic distributions, making it difficult to compute analytically. To this end, we use the reparameterization trick~\cite{kingma2014auto}, which allows us to reform samples from $p(Z^m|E^m)$ by a bivariate deterministic transformation ($\mu^m$ and $\sigma^m$), separating stochastic noise ($\epsilon^m$) from the parameters of the distribution. Thus it facilitates gradient-based optimization through standard back-propagation. Finally, the representations $Z^m$ are sampled using the reparameterization trick, formulated as:\begin{equation}
  Z^m([\mu^m,\sigma^m],\epsilon^m)=\mu^m+\sigma^m\odot \epsilon^m,
  \label{eq:reparameterization trick}
\end{equation}
where $\epsilon^m \sim \mathcal{N}(0,1)$ is a random vector the same size as $Z^m$, $\odot$ is the operator for elementwise multiplication. The resulting $Z^m$ are then used as stochastic inputs in Eq.~\ref{eq:LRIB' unidef}. From a denoising perspective, the above Eq.~\ref{eq:reparameterization trick} highlights that the hidden representations are perturbed by self-adaptive Gaussian noise during training, regulated by the standard deviation $\sigma^m$. Unlike deterministic noise injection in the feature space, this approach was shown in prior work to enhance robustness against noise during the test phase~\cite{li2017collaborative}.

We proceed to focus on calculating LRIB principle in Eq.~\ref{eq:LRIB' unidef}. For the first term $I_{\alpha}^k(E^m_A;Z^m_A)$, the estimation is highly challenging or even infeasible, particularly when dealing with high-dimensional distributions commonly encountered in deep learning. To address this issue, we employ low-rank R\'enyi's entropy measure to directly optimize this term using Eq.~\ref{eq:mutual} because of the advantages it brings. For the second term $I(Z^m;Y^m)$, we employ variational approximations~\cite{blei2017variational} because this term is equivalent to the cross-entropy (CE) loss or Mean Absolute Error (MAE) loss for classification tasks or regression tasks, respectively. Recall that calculating this term needs to directly calculate and optimize $p(y^m|z^m)$ according to Eq.~\ref{eq:IZY}. However, the optimization is intractable as the decoder distribution $p(y^m|z^m)$ can take the form of any valid conditional distributions and most of which are not even differentiable. To solve this problem, the variational method offers a practical solution by assuming that the decoder belongs to a tractable family of distributions $Q$ and finding a distribution $q(y^m|z^m)$ in that family that is closest to the optimal distribution of the decoder measured by the KL-divergence. In this context, $q(y^m|z^m)$ serves as the variational approximation to $p(y^m|z^m)$. Based on the property of KL-divergence being non-negative $\mathrm{KL}[p(y^m|z^m)\|q(y^m|z^m)] \geq 0$, we can obtain the inequality:
\begin{equation}
  \begin{aligned}
    \int p(y^m|z^m) \log p(y^m|z^m) \mathrm{d} y \geq \int p(y^m|z^m) \log q(y^m|z^m) \mathrm{d} y.
  \end{aligned}
\end{equation}

Furthermore, we can obtain the lower bound through variational distribution:
\begin{eqnarray}
  \begin{aligned}
     I(Y^m;Z^m) =& \int p(y^m, z^m) \log \frac{p(y^m | z^m)}{p(y^m)} \mathrm{d} y^m \mathrm{~d} z^m\\
     \geq & \int p(y^m, z^m) \log \frac{q(y^m | z^m)}{p(y^m)} \mathrm{d} y^m \mathrm{~d} z^m \\
     \geq & \int p(y^m, z^m) \log q(y^m | z^m) \mathrm{d} y^m \mathrm{~d} z^m
  \end{aligned} \label{eq:lower bound of IZY}
\end{eqnarray}
 The whole proof can be found in~\cite{blei2017variational}.

The selection of the tractable family $Q$ for the decoder distributions $q(y^m|z^m)$ can be tailored to the specific MSA prediction task. In our paper, we consider both classification and regression tasks. For the classification task, the family $Q$ of the decoder distributions $q(y^m|z^m)$ can be chosen as an MLP where the output is squashed through a Sigmoid function. The computation of $\log q(y^m|z^m)$ is shown as follows:
\begin{eqnarray} \begin{aligned}
    \hat{y}^m=&\text{Sigmoid}(\text{MLP}(z^m)), \\
    q(y^m|z^m)\cong & q(y|z^m) = (\hat{y}^m)^y \cdot(1-\hat{y}^m)^{1-y}, \\
    \log q(y|z^m)=&y \log \hat{y}^m+(1-y) \log (1-\hat{y}^m),
  \end{aligned}\label{eq:classification decoder}
\end{eqnarray}
where $\text{Sigmoid}(x)=1/(1+e^{-x})$. We use the overall true label $y$ to guide the learning of unimodal representation so that $y^m\cong y$. It is easy to observe that maximizing the term $\log q(y^m|z^m)$ is equivalent to the minimization of the CE loss between unimodal prediction output $\hat{y}^m$ and the groundtruth label $y$.

For regression task, we calculate $\log q(y^m|z^m)$ as:
\begin{equation}
  \begin{aligned} 
  o^m =& \text{MLP}(z^m),\\
  q(y^m | z^m)\cong & q(y|z^m)=e^{-\left |y-o^m\right |+C}, \\ 
  \log q(y | z^m)=&-\left |y-o^m\right |+C,
  \end{aligned} \label{eq:regression decoder}
\end{equation}
where $C$ is a constant and the target $y$ is a continuous variable. We can also observe that maximization of $\log q(y^m|z^m)$ is equivalent to the minimization of the MAE loss between the unimodal representative output $o^m$ and the target $y$ .

\subsection{Attention Bottleneck Fusion}
\label{sec:4.3}
\begin{figure}
\centering
\includegraphics[width=\linewidth]{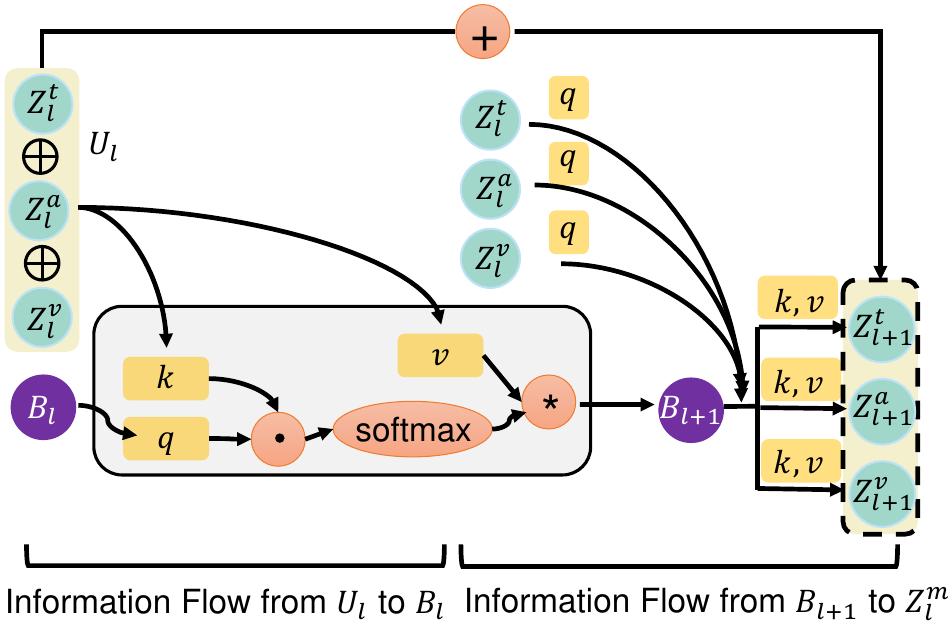}
\caption{Attention bottleneck fusion module.The process enables iterative information flow, where cross-modal information is first aggregated into bottleneck embeddings, and then redistributed to enhance modality-specific representations.}
\label{fig:attention-fusion}
\end{figure}

Inspired by the concept of bottleneck tokens introduced in~\cite{nagrani2021attention}, we propose a novel attention bottleneck fusion module that incorporates a lightweight attention mechanism with learnable bottleneck vectors. The goal is to suppress redundant information transmission and facilitate efficient multimodal interaction. To illustrate in this section, we consider the tri-modal case as an example. As depicted in Figure~\ref{fig:attention-fusion}, instead of allowing direct attention-based interactions between modalities which can be computationally expensive and noisy, our module employs a shared compact bottleneck as an intermediary, significantly reducing attention computation and improving feature selectivity.

The fusion module takes modality-specific feature sequences $Z^t, Z^a, Z^v$ as input, which have been processed under the LRIB constraint. These sequences are concatenated to form a unified representation $U_0 \in \mathbb{R}^{(l_t+l_a+l_v) \times d_m}$:
\begin{equation}
  \begin{aligned}
    U_0=Concat(Z^t, Z^a, Z^v).
  \end{aligned}
\end{equation}

We introduce a set of learnable bottleneck embeddings $B \in \mathbb{R}^{l_b \times d_m}$ ($l_b \ll l_m$) to mediate information exchange across modalities. The limited embedding capacity encourages selective information transfer. Cross-modal attention is applied between the unified representation $U$ and the bottleneck embeddings $B$, which models dependencies between a source sequence and a target sequence. Given source $X_s$ and target $X_t$, queries, keys, and values are computed as $Q_t=X_tW_{Q_t}$, $K_s=X_sW_{K_s}$ and $V_s=X_sW_{V_s}$, respectively. The weights $W_{Q_t} \in \mathbb{R}^{d_t \times d_k}$, $W_{K_s} \in \mathbb{R}^{d_s \times d_k}$ and $W_{V_s} \in \mathbb{R}^{d_s \times d_v}$ are learnable projection matrices. In this context, one single head of the first layer of attention bottleneck fusion operation can be formulated as follows:
\begin{equation}
  \begin{aligned}
    B_1=&\text{CM}_{s\to t}(U_0,B_0)\\
    =&\operatorname{softmax}\left(\frac{B_0U_0^T}{\sqrt{d_{U_{0}}}}\right)U_0,\\
    =&\operatorname{softmax}\left(\frac{B_0 W_{Q_{B_0}} W_{K_{U_0}}^T U_0^T}{\sqrt{d_{U_0}}}\right) U_0 W_{V_{U_0}},
  \end{aligned}\label{eq:Bflow2Z}
\end{equation}
here our source is $U_0$ with $d_{U_0}$ representing its dimensionality and target sequence is a randomly initialized embedding $B_0$. The result $B_1$ represents the updated bottleneck embeddings enriched with cross-modal context.

The fusion is performed iteratively over $M$ layers. At each layer $l$, the bottleneck embeddings first aggregate global information from the current unified representation $U_l$:
\begin{equation}
    \begin{aligned}
        B_{l+1}&=\text{CM}_{s\to t}(U_l,B_l).
    \end{aligned}\label{eq:Bflow2Z}
\end{equation}
Subsequently, each modality $Z_{l}^m$ interacts with the updated bottleneck $B_{l+1}$ to incorporate shared cross-modal information, with the update rule defined as:
\begin{equation}
  \begin{aligned}
    Z_{l+1}^m&=Z_{l}^m+\gamma^m \text{CM}_{s\to t}(B_{l+1},Z_{l}^m),
  \end{aligned}\label{eq:Zflow2B}
\end{equation}
where $\gamma^m$ are the regularization coefficients of the attention mechanism for each modality. It is noteworthy that the bottleneck embeddings serve different roles in Eq.~\ref{eq:Bflow2Z} and Eq.~\ref{eq:Zflow2B}, representing different information exchange directions between a "bottleneck" and unimodal information: the shared bottleneck embeddings absorbing multimodal context from all modalities, and selectively distributing it back to each modality. After $M$ layers of fusion, we obtain the final modality-enhanced embeddings $Z_{M}^t, Z_{M}^a, Z_{M}^v$, each infused with context-aware, cross-modal information mediated by the attention bottleneck.


\begin{equation}
  \begin{aligned}
    Z=\mathrm{ReLU}(Z^{t}_M),
  \end{aligned}
\end{equation}
and apply the refined LRIB objective to guide the multimodal representation learning, yielding the desirable representation $\Tilde{Z}$ :
\begin{eqnarray}
  \begin{aligned}
\mathcal{L}_{Multi-LRIB}=& I_{\alpha}^k(Z_A; \Tilde{Z}_A)-\beta I(\Tilde{Z};Y),\\
\end{aligned}\label{eq:multi-LR}
\end{eqnarray}
 where $Z_A, \Tilde{Z}_A$ are square matrices acting on $Z$ and $\Tilde{Z}$, respectively. Similarly, $\Tilde{Z}$ is obtained through a multimodal variational encoder $\phi$:
 \begin{equation}
 \begin{aligned}
   \left[\mu, \log \sigma\right]=&\operatorname{ReLU(}\operatorname{MLP}\left(Z)\right),\\
  \Tilde{Z}([\mu,\sigma],\epsilon)=&\mu+\sigma\odot \epsilon.
 \end{aligned}
\end{equation}
 As mentioned in Section~\ref{sec:Unimodal Representation Learning}, maximizing the latter term $I(\Tilde{Z};Y)$ in Eq.~\ref{eq:multi-LR} boils down to minimizing cross-entropy in classification tasks or Mean Absolute Error for regression tasks ~\citep{achille2018information,amjad2019learning}.
\begin{equation}
\begin{aligned}
  &\log q(y|z)=y \log \hat{y}+(1-y) \log (1-\hat{y}), \,\,\, \text{for classification} \\
   &\log q(y | z)=-\left |y-o\right |+C, \,\,\, \text{for regression}
\end{aligned}
\end{equation}
where $\hat{y}=\text{Sigmoid}(\text{MLP}(\Tilde{z}))$ for classification task,$o = \text{MLP}(\Tilde{z})$ for regression task, representing the outputs of the final multimodal representation. $y$ is the groundtruth label.

\begin{algorithm}[!t]\label{alg:1}
\small{
	\caption{DIB with Bottleneck Fusion.}
	\LinesNumbered
	\KwIn{Sample pairs $\{E^{m},Y\}$, $m\in \mathcal{M}$, the kernel width $\tau$, $\alpha > 0$, $k$, $\beta$.}
	\KwOut{Prediction $\hat{Y}$.}
    \While{not converged}{
      \texttt{// Unimodal Learning} \\
  	\For{$m\in \mathcal{M}$}{
  		$Z^m \gets \phi_m(E^m)$ \\
  		\texttt{//} Calculate pairwise distance \\
        $D_{Z^m} \gets {L_2}\_\text{distance}(Z^m)$ \\
        $D_{E^m} \gets {L_2}\_\text{distance}(E^m)$ \\
        $Z^m_A, E^m_A \gets \exp(-D_{Z^m} \!/\! \tau_{Z^m}^2), \exp(-D_{E^m} \!/\! \tau_{E^m}^2)$ \\
        $Z^m_A, E^m_A \gets Z^m_A / tr(Z^m_A), E^m_A / tr(E^m_A) $ \\
        Calculate eigenvalues of $Z^m_A, E^m_A$ \\
        Calculate $I_{\alpha}^k(E^m_A;Z^m_A)$ \\
        Calculate the lower bound of $I(Y^m;Z^m)$ as in Eq. (\ref{eq:lower bound of IZY}) \\
  	}
      Calculate $\mathcal{L}_{Uni-LRIB}$ as in Eq. (\ref{eq:LRIB' unidef}) \\
      \texttt{// Multimodal Learning} \\
      $Z \gets BottleneckFusion(Z^m)$ \\
      $\Tilde{Z} \gets \phi(Z)$ \\
      Calculate $\mathcal{L}_{Multi-LRIB}$ as in Eq. (\ref{eq:multi-LR}) \\
      \texttt{// Joint Optimization} \\
      $\hat{Y} \gets \psi(\Tilde{Z})$ \\
      $\mathcal{L}_{LRIB}=\mathcal{L}_{Uni-LRIB}+\mathcal{L}_{Multi-LRIB}$ as in Eq. (\ref{eq:Total LRIB Loss}) \\
  	Update model parameters $\Phi$ 
    }
    \Return{$\hat{Y}$.}}
\end{algorithm}

\begin{figure}
\centering
\includegraphics[width=\linewidth]{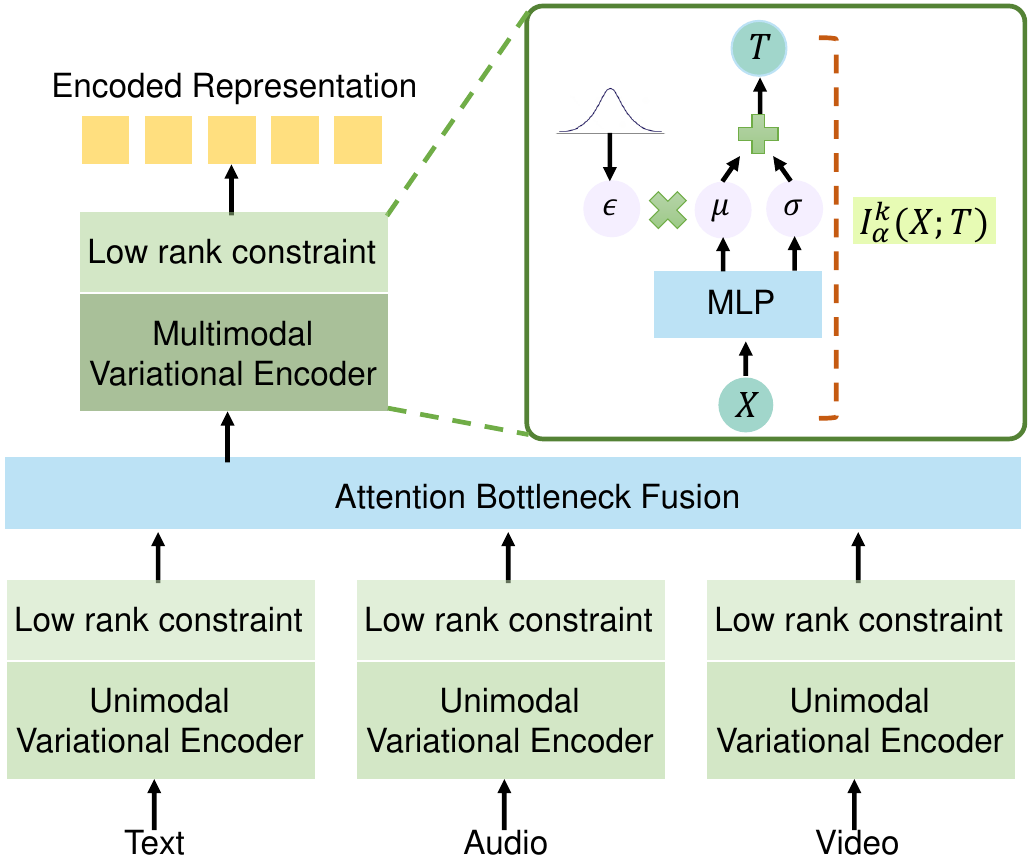}
\caption{Joint encoding of unimodal and multimodal representations where the representations are obtained through variational encoders and optimized using the low-rank R\'enyi's entropy training objective.}\label{fig:encoder}
\end{figure}

\begin{figure}
\centering
\includegraphics[width=\linewidth]{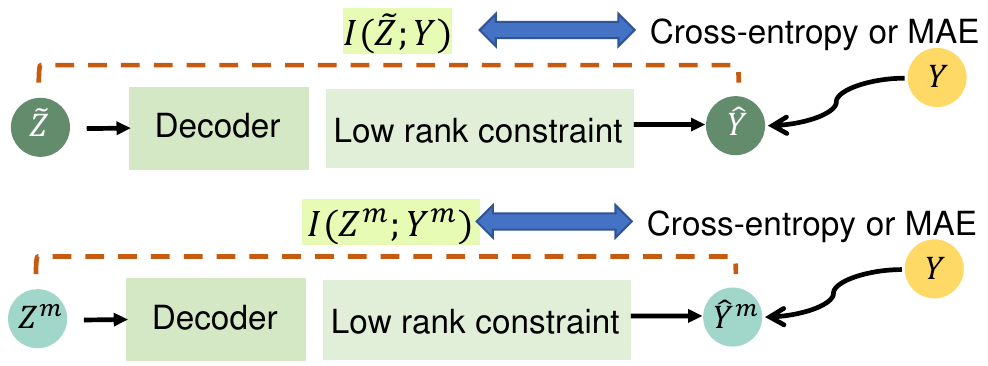}
\caption{The decoded multimodal and unimodal representations are used to compute the low-rank R\'enyi’s entropy, which corresponds to the loss computed with respect to the overall label.}\label{fig:decoder}
\end{figure}

\subsection{LRIB for Multimodal Representation Learning}
\label{sec:4.5}
After the fusion process, we adopt the representation of the textual modality $Z^{t}_M$ as the final predictive embedding for downstream tasks, due to the following reasons: 1) the textual modality typically enjoys rich and explicit contextual information used for accurate sentiment interpretation, and when processed by advanced pre-trained language models such as BERT, its representation is strengthened, providing a more powerful and nuanced understanding compared to other modalities, as demonstrated in \citep{hazarika2020misa, yang2022multimodal}. 2) non-textual modalities, e.g., audio and video inputs, are more susceptible to disruptions such as lighting conditions, motion blur, and background interference, resulting in degradation performance. Moreover, empirical evidence from ablation studies in Section \ref{sec:ablation} further corroborates the effectiveness of prioritizing the textual modality as the dominant modality within our framework. Therefore, leveraging the representation of textual modality enriched with complementary information from other modalities is sufficient to achieve strong generalization performance.

We then use ReLU activation to output the feature representation:

\subsection{Joint Optimization}
The joint training process involving encoder and decoder part is shown in Figure\ref{fig:encoder} and Figure\ref{fig:decoder}. By applying the LRIB principle at both the unimodal and multimodal levels, we ensure that each modality is individually optimized to be informative yet compact, and that the combined multimodal representation $\Tilde{Z}$ captures the most relevant information for the task without redundancy or noise. 

At last, the task-specific outputs are obtained by decoding the unified multimodal representation $\Tilde{Z}$. We optimize the proposed DIB model through the overall LRIB loss:
\begin{equation}\label{eq:Total LRIB Loss}
  \begin{aligned}
    \mathcal{L}_{LRIB}=\mathcal{L}_{Uni-LRIB}+\mathcal{L}_{Multi-LRIB},
  \end{aligned}
\end{equation}
which involves joint optimization objectives over both unimodal and multimodal representation learning.

The overall framework of the proposed method is presented in Algorithm~\ref{alg:1}. Without further specifications, the indivisible kernel employed in our algorithm is the Gaussian kernel.

\section{Experiment Settings}
\subsection{Dataset}
We adopt four widely-used datasets for MSA: CMU-MOSI~\cite{zadeh2016mosi}, CMU-MOSEI~\cite{zadeh2018multimodal}, CH-SMIS~\cite{yu2020ch} and MVSA-Single~\cite{niu2016sentiment}. The first three datasets include visual, audio, and textual modalities, whereas MVSA-Single consists of paired image-text data. Detailed dataset statistics are presented in Table~\ref{dataset_split}. Following common practice, we use the unaligned versions of CMU-MOSI, CMU-MOSEI, and CH-SIMS. Serving as standard benchmarks, these datasets provide a diverse testbed for assessing generalization. CMU-MOSI and CMU-MOSEI include spontaneous tri-modal English content, while CH-SIMS introduces Chinese data for cross-lingual evaluation. MVSA-Single adds image-text pairs from social media, providing a more practical assessment.

\textbf{MOSI.} The dataset, developed in English, consists of 2,199 video segments, each representing an utterance from 93 videos, featuring 89 distinct narrators discussing various topics. Sentiment values range from -3 (strongly negative) to +3 (strongly positive), representing both polarity and relative strength of expressed sentiment. The dataset is divided into 1,284 training samples, 229 validation samples, and 686 test samples, respectively.

\textbf{MOSEI.} The dataset is an extension of CMU-MOSI. It contains 23,454 video clips collected from YouTube, encompassing diverse factors such as spontaneous expressions, head poses, occlusions, and varying lighting conditions. Sentiment values range from -3 (strongly negative) to +3 (strongly positive), as well as emotion labels across six categories: anger, disgust, fear, happiness, sadness, and surprise. It is partitioned into 16,326 training instances, 1,871 validation instances, and 4,659 test instances.

\textbf{CH-SIMS.} The dataset is a Chinese multimodal sentiment dataset comprising 2,281 video clips from various sources such as movies and TV shows. It includes diverse expressions and head poses. The dataset is split into 1,368 training samples, 456 validation samples, and 457 test samples, each manually labeled with a sentiment score from -1 (negative) to +1 (positive).

\textbf{MVSA-Single.} MVSA-Single is a popular image-text sentiment dataset crawled from Twitter, consisting of 5129 image-text pairs. Each pair is annotated with a human-labeled sentiment category: positive, negative, neutral.

\begin{table}[h]
\caption{Statistics of the adopted dataset.}\label{dataset_split}
\centering
\setlength{\tabcolsep}{6pt} 
\renewcommand{\arraystretch}{1.1} 
\begin{tabular}{lcccc}
\toprule
Dataset & Train & Validation & Test & Total \\ \midrule
\textbf{CMU-MOSI} & 1284 & 229 & 686 & 2199 \\ 
\textbf{CMU-MOSEI} & 16326 & 1871 & 4659 & 22856 \\ 
\textbf{CH-SIMS} & 1368 & 456 & 457 & 2281 \\ 
\textbf{MVSA-Single} & 3608 & 451 & 452 & 4511 \\ 
\bottomrule
\end{tabular}
\end{table}

\subsection{Baselines}
\label{sec:5.2}
To comprehensively evaluate the performance of our proposed DIB method on MSA task, we select both trimodal and bimodal baselines.

\textbf{Trimodal Baselines.} The compared baselines for trimodal datasets (visual-audio-text modality) include: \, \, \, \, \, 1)\textbf{Graph-MFN}~\cite{zadeh2018multimodal} employs a fusion graph approach to model unimodal, bimodal, trimodal interactions and enhances the interpretability by focusing on the dynamics of contextual information. 2) \textbf{MulT}~\cite{tsai2019multimodal} introduces the multimodal Transformer which applies the attention mechanism to capture intricate cross-modal interactions and alignments. 3) \textbf{GraphCAGE}~\cite{wu2021graph} adapts capsule networks and graph convolutional networks to handle long unaligned sequences, capturing long-range dependencies of multimodal information. 4) \textbf{TFR-Net}~\cite{yuan2021transformer} addresses modality incompleteness through the Transformer-based feature reconstruction module. 5)\textbf{ MMIM}~\cite{han2021improving} extracts task-relevant information by maximizing mutual information between fusion representation and unimodal input. 6) \textbf{Self-MM}~\cite{yu2021learning} leverages self-supervised learning to generate unimodal labels and jointly trains unimodal tasks and multimodal task to learn both consistency and differences across modalities. 7) \textbf{HyCon}~\cite{mai2022hybrid} captures intra-/inter-modal dynamics and inter-class relationships by designing different contrastive losses. 8) \textbf{SUGRM}~\cite{hwang2023self} designs a simpler calculation of unimodal labels using recalibrated features. 9) \textbf{MIB}~\cite{mai2022multimodal} employs the general IB principle to learn the minimal sufficient multimodal representation. 10) \textbf{PS-Mixer}~\cite{lin2023ps} introduces a polar-vector mechanism to identify sentiment polarity and develops the MLP-communication module to minimize noise interference while enhancing multimodal interactions. 11) \textbf{EMT}~\cite{sun2023efficient} integrates utterance-level representations from multiple modalities to create a global multimodal context that dynamically enhances local unimodal features for mutual improvement. 12) \textbf{ALMT}~\cite{zhang2023learning} considers the language modality as the dominant feature, leveraging it to identify and address the irrelevant and conflicting information in visual and auditory features. 13) \textbf{TMBL}~\cite{huang2024tmbl} integrates bimodal and trimodal binding mechanisms with fine-grained convolutional modules and employs both similarity and dissimilarity losses to enhance model convergence. 14) \textbf{MFN}~\cite{zadeh2018memory} maintains a temporal memory of modality interactions, allowing for a more dynamic and context-aware fusion. 15) \textbf{MUG}~\cite{mai2024meta} proposes a meta-learning framework for weakly supervised unimodal label learning, using denoising tasks and bi-level optimization to enhance multimodal inference.

\textbf{Bimodal Baselines.} In addition to trimodal evaluation, we also include experiments in the bimodal setting (image-text modality) using the MVSA-Single dataset. The compared baselines include: 1) \textbf{HSAN} \cite{xu2017analyzing} applies a hierarchical semantic attentional network for multimodal sentiment analysis. 2) \textbf{CoMN-Hop6} \cite{xu2018co} develops a stacked co-memory network to iteratively model the interactions between modalities. 3) \textbf{MVAN-M} \cite{yang2020image} captures and integrates cross-modal dependencies through a multi-view attentional framework. 4) \textbf{MGNNS} \cite{yang2021multimodal} utilizes a multi-channel graph neural network with sentiment-awareness to capture the global co-occurrence characteristics in texts and images. 5) \textbf{CLMIF} \cite{li2022clmlf} employs a Transformer encoder for token-level modality alignment and introduces dual contrastive learning tasks to enhance the learning of shared sentiment features. 6) \textbf{VSA-PF}~\cite{chen2024holistic} leverages complementary strengths of diverse pre-trained models to enhance robustness. 7) \textbf{CIGNN}~\cite{wang2024multimodal} proposes a cross-instance graph neural network that captures global co-occurrence patterns across image-text pairs.

\subsection{Evaluation Metrics}
For CMU-MOSI and CMU-MOSEI datasets, following previous works~\citep{hazarika2020misa,zhang2023learning}, we employ Weighted F1-score (F1), seven-class accuracy (ACC-7), and binary accuracy (ACC-2) to evaluate the overall performance of the models on classification tasks. The Mean Absolute Error (MAE) and Pearson correlation (Corr) are used to measure the performance on regression tasks. For CH-SIMS dataset, in line with previous works~\citep{mai2024meta,hwang2023self}, we report ACC-2, F1, MAE, Corr metrics. For MVSA-Single dataset, due to its highly unbalanced label distribution, we follow prior work~\citep{yang2021multimodal} and report ACC-2 and Weighted F1-score instead of the Micro F1 used in the original paper, as this provides a more realistic evaluation. Note that the higher the metric value, the better the performance, except for MAE. 

To assess the performance of the models in noisy settings, we calculate the performance decline for each metric:
\begin{equation}
  \begin{aligned}
    \mathrm{Decline(\%)}=\frac{M_{old}-M_{new}}{M_{old}} \times 100\%,
  \end{aligned}
\end{equation}
where $M_{old}$ and $M_{new}$ represent the prediction performance metrics over the original data and noise data, respectively.

\subsection{Implementation Details}
\label{sec:Implementation Details}
For MOSI, MOSEI, CH-SIMS dataset, we employ the pre-trained BERT-base model to extract text embeddings, and use Transformer modules with 5 attention heads for extracting video and audio embeddings, with dimensions of 768 (text), 74 (video), and 47 (audio), respectively. To align the heterogeneous embeddings from different modalities, we apply modality-specific MLP layers to project them to a common hidden dimension of 50. For MVSA-Single dataset, we use the ViT-B/16 model and Transformer encoders with 8 attention heads to extract both visual and textual embeddings, each projected to a unified 512-dimensional space. 
The number of Transformer layers and Attention Bottleneck Fusion layers is set to 3 across all datasets. We adopt the Adam optimizer and perform grid search for the learning rate from \{1e-5, 2e-5, 6e-5, 1e-4, 3e-4\}, and for the information bottleneck-related coefficients, we perform a grid search over $\alpha$ in the range of $[1.1, 2.0]$ with a step size of $0.1$, and $\beta \in \{1e-6, 1e-5, 1e-4, 1e-2\}$ for both unimodal and multimodal branches. The optimal values of the learning rate, $\alpha$ and $\beta$ selected for each dataset are summarized in Table \ref{tab:hyperparams}.
Batch sizes are set to 32, 128, and 32 for CMU-MOSI, CMU-MOSEI, and CH-SIMS respectively, while for MVSA-Single, a smaller batch size of 16 is used due to its larger input dimensionality. The dropout rate is uniformly set to 0.5 for all datasets to mitigate overfitting. All experiments are conducted with 5 different random seeds to ensure robustness, and we adopt the official dataset splits provided by the original papers to guarantee a fair comparison across baselines.
We train all models for 50 epochs using Python 3.9.18, PyTorch 2.2.2, and CUDA 12.2 on NVIDIA RTX 4090 GPUs. Implementation details are provided in Table \ref{tab:hyperparams}.

For low rank matrix-based R\'enyi’s entropy estimation, we adopt a Gaussian kernel and dynamically estimate the kernel bandwidth $\sigma^2$ within each training batch. Specifically, $\sigma^2$ is determined by the average of the top-5 nearest pairwise Euclidean distances among samples, which enables adaptive scaling across datasets of different feature magnitudes. The resulting Gram matrix is trace-normalized to ensure it is a valid positive semi-definite matrix with unit trace. For eigenspectrum truncation, we employ a fixed-rank strategy with truncation rank $k=10$, chosen to balance information preservation and computational cost. The R\'enyi’s order is set to $\alpha=1.9$, which yields the best overall performance across datasets. All these settings follow established practices in matrix-based R\'enyi’s entropy estimation~\cite{giraldo2014measures} and were found to provide stable and reproducible results.

\begin{table}[h]
\centering
\setlength{\tabcolsep}{4.3pt} 
\renewcommand{\arraystretch}{1.1} 
\caption{Implementation details of DIB for each dataset.}
\begin{tabular}{lcccc}
\toprule
\textbf{Parameter} & \textbf{MOSI} & \textbf{MOSEI} & \textbf{CH-SIMS} & \textbf{MVSA} \\
\midrule
\makecell[l]{1) Learning rate of\\ language encoder} & 1e-5 & 1e-5 & 1e-5 & 1e-5 \\
\makecell[l]{2) Learning rate of\\ the whole model} & 2e-5 & 2e-5 & 2e-5 & 2e-5 \\
3) Batch size        & 32      & 128      & 32      & 16     \\
4) Dropout rate      & 0.5     & 0.5     & 0.5     & 0.5     \\
5) Hidden dimension  & 50     & 50     & 50     & 512    \\
6) $\alpha$         & 1.9    & 1.9     & 1.9 & 1.9 \\
\makecell[l]{7) $\beta$ of unimodal\\ learning}         & 1e-5    & 1e-5     & 1e-5 & 1e-5 \\
\makecell[l]{8) $\beta$ of multimodal\\ learning}         & 1e-5    & 1e-4     & 1e-5 & 1e-5 \\
\makecell[l]{9) Number of \\attention heads}         & 5    & 5     & 5 & 8 \\
\bottomrule
\end{tabular}
\label{tab:hyperparams}
\end{table}

\begin{table*}[!ht]
\caption{Performance comparison on the CMU-MOSI and CMU-MOSEI benchmarks. $\dagger$: results from~\cite{mai2024meta}. $\ddagger$: results from~\cite{sun2023efficient}. $*$: reproduced using publicly available source codes and original hyper-parameters under the same setting. We run each model five times and report average results. When calculating ACC-2 and F1 score, we exclude the neutral utterances.}
\setlength{\tabcolsep}{6.5pt} 
\renewcommand{\arraystretch}{0.4} 
\centering
\renewcommand{\arraystretch}{0.78}
\small{
\begin{tabularx}{\linewidth}{lcccccccccc} 
  \toprule[1.2pt]
  \multirow{2.5}{*}{\textbf{Models}} & \multicolumn{5}{c}{\textbf{CMU-MOSI}} & \multicolumn{5}{c}{\textbf{CMU-MOSEI}} \\
  \cmidrule(lr){2-6} \cmidrule(lr){7-11}
  & \footnotesize\textbf{ACC-7}$\uparrow$ & \footnotesize\textbf{ACC-2}$\uparrow$ & \footnotesize\textbf{F1}$\uparrow$ & \footnotesize\textbf{Corr}$\uparrow$ & \footnotesize\textbf{MAE}$\downarrow$ & \footnotesize\textbf{ACC-7}$\uparrow$ & \footnotesize\textbf{ACC-2}$\uparrow$ & \footnotesize\textbf{F1}$\uparrow$ & \footnotesize\textbf{Corr}$\uparrow$ & \footnotesize\textbf{MAE}$\downarrow$ \\
  \midrule
  Graph-MFN$^{\dagger}$~\cite{zadeh2018multimodal} & 34.4 & 80.2 & 80.1 & 0.656 & 0.939 & 51.9 & 84.0 & 83.8 & 0.725 & 0.569 \\
  MulT$^{\ddagger}$~\cite{tsai2019multimodal} & 40.4 & 83.4 & 83.5 & 0.725 & 0.846 & 52.6 & 83.5 & 83.6 & 0.731 & 0.564 \\ 
  GraphCAGE~\cite{wu2021graph} & 35.4 & 82.1 & 82.1 & 0.684 & 0.933 & 48.9 & 81.7 & 81.8 & 0.670 & 0.609 \\ 
  TFR-Net$^{\ddagger}$~\cite{yuan2021transformer} & 46.1 & 84.0 & 84.0 & 0.789 & 0.721 & 52.3 & 83.5 & 83.8 & 0.756 & 0.551 \\ 
  MMIM$^{\dagger}$~\cite{han2021improving} & 45.0 & 85.1 & 85.0 & 0.781 & 0.738 & 53.1 & 85.1 & 85.0 & 0.752 & 0.547 \\ 
  Self-MM$^{\dagger}$~\cite{yu2021learning} & 45.8 & 84.9 & 84.8 & 0.785 & 0.731 & 53.0 & 85.2 & 85.2 & 0.763 & 0.540 \\ 
  HyCon$^{\dagger}$~\cite{mai2022hybrid} & 46.6 & \underline{85.2} & \underline{85.1} & 0.779 & 0.741 & 52.8 & 85.4 & 85.6 & 0.751 & 0.554 \\
  SUGRM$^{\dagger}$~\cite{hwang2023self} & 44.9 & 84.6 & 84.6 & 0.772 & 0.739 & \textbf{53.7} & 85.4 & 85.3 & 0.759 & 0.537 \\ 
  \midrule
  \multirow{2}{*}{MIB$^*$~\cite{mai2022multimodal}} & \underline{46.8} & 85.1 & \underline{85.1} & \underline{0.795} & 0.728 & 52.9 & 84.4 & 84.4 & \underline{0.786} & 0.592 \\
  & {\scriptsize ± 0.11} &{\scriptsize ± 0.20} & {\scriptsize ± 0.20} & {\scriptsize ± 0.0010}& {\scriptsize ± 0.0100}& {\scriptsize ± 0.40}& {\scriptsize ± 0.40}& {\scriptsize ± 0.30}& {\scriptsize ± 0.0010}& {\scriptsize ± 0.0020}\\
  
  \multirow{2}{*}{PS-Mixer$^*$~\cite{lin2023ps}} & 41.0 & 82.2 & 82.1 & 0.772 & 0.795 & 52.8 & \textbf{86.1} & \textbf{86.2} & 0.767 & 0.537 \\
  & {\scriptsize ± 0.42} &{\scriptsize ± 0.24} & {\scriptsize ± 0.16} & {\scriptsize ± 0.0510}& {\scriptsize ± 0.0157}& {\scriptsize ± 0.16}& {\scriptsize ± 0.50}& {\scriptsize ± 0.01}& {\scriptsize ± 0.0008}& {\scriptsize ± 0.0014}\\
  
  \multirow{2}{*}{EMT$^*$~\cite{sun2023efficient}} & \underline{46.8} & 85.1 & \underline{85.1} & 0.794 & \textbf{0.713} & 53.1 & 85.7 & 85.7 & 0.774 & \underline{0.534} \\
  & {\scriptsize ± 0.67} &{\scriptsize ± 0.43} & {\scriptsize ± 0.43} & {\scriptsize ± 0.0012}& {\scriptsize ± 0.0066}& {\scriptsize ± 0.42}& {\scriptsize ± 0.10}& {\scriptsize ± 0.15}& {\scriptsize ± 0.0010}& {\scriptsize ± 0.0012}\\
  
  \multirow{2}{*}{ALMT$^*$~\cite{zhang2023learning}} & 45.3 & 85.1 & \underline{85.1} & 0.793 & 0.721 & 53.0 & 85.7 & 85.7 & 0.779 & \textbf{0.527} \\
  & {\scriptsize ± 0.10} &{\scriptsize ± 0.22} & {\scriptsize ± 0.38} & {\scriptsize ± 0.0057}& {\scriptsize ± 0.0244}& {\scriptsize ± 0.13}& {\scriptsize ± 0.08}& {\scriptsize ± 0.09}& {\scriptsize ± 0.0010}& {\scriptsize ± 0.0012}\\
  
  \multirow{2}{*}{TMBL$^*$~\cite{huang2024tmbl}} & 38.3 & 83.3 & 83.4 & 0.724 & 0.869 & 48.7 & 85.5 & 85.5 & 0.751 & 0.602 \\
  & {\scriptsize ± 0.56} &{\scriptsize ± 0.29} & {\scriptsize ± 0.40} & {\scriptsize ± 0.0014}& {\scriptsize ± 0.0025}& {\scriptsize ± 0.17}& {\scriptsize ± 0.15}& {\scriptsize ± 0.15}& {\scriptsize ± 0.0005}& {\scriptsize ± 0.0078}\\
  \midrule
  \multirow{2}{*}{\textbf{Ours (DIB)}} & \textbf{47.4} & \textbf{85.6} & \textbf{85.6} & \textbf{0.800} & \underline{0.715} & \underline{53.5} & \underline{86.0} & \underline{86.0} & \textbf{0.790} & 0.588 \\
  & {\scriptsize ± 0.13} &{\scriptsize ± 0.09} & {\scriptsize ± 0.09} & {\scriptsize ± 0.0010}& {\scriptsize ± 0.0060}& {\scriptsize ± 0.20}& {\scriptsize ± 0.11}& {\scriptsize ± 0.05}& {\scriptsize ± 0.0020}& {\scriptsize ± 0.0009}\\
  \bottomrule[1.2pt]
\end{tabularx}}
\label{tab:mosi&mosei}
\end{table*}

\begin{table}[h]
\caption{Performance comparison on the CH-SIMS benchmark. $\dagger$: results from~\cite{mai2024meta}. $\ddagger$: results from~\cite{sun2023efficient}.}
\centering
\renewcommand{\arraystretch}{0.9}
\setlength{\tabcolsep}{6pt} 
\begin{tabular}{lcccc}
\toprule
\textbf{Models} & \textbf{MAE $\downarrow$} & \textbf{Corr $\uparrow$} & \textbf{ACC-2 $\uparrow$} & \textbf{F1 $\uparrow$} \\
\midrule
MFN$^{\dagger}$ ~\cite{zadeh2018memory}      & 0.435   & 0.582   & 77.90   & 77.88 \\
Graph-MFN$^{\dagger}$~\cite{zadeh2018multimodal}    & 0.445   & 0.578   & 78.77   & 78.21 \\
MulT$^{\dagger}$~\cite{tsai2019multimodal}      & 0.453   & 0.564   & 78.56   & 79.66 \\
TFR-Net$^{\ddagger}$~\cite{yuan2021transformer}     & 0.437   & 0.583   & 78.00   & 78.10 \\
MMIM$^{\ddagger}$~\cite{han2021improving}       & 0.422   & 0.597   & 78.30   & 78.20 \\
Self-MM$^{\dagger}$~\cite{yu2021learning}     & 0.425   & 0.592   & 80.04   & \underline{80.44} \\
SUGRM$^{\dagger}$~\cite{hwang2023self}      & 0.418   & 0.596   & 79.26   & 79.13 \\
EMT$^\ddagger$~\cite{sun2023efficient}       & \textbf{0.396}   & \underline{0.623}   & 80.10   & 80.10 \\
MUG$^{\dagger}$ ~\cite{mai2024meta}      & \underline{0.415}   & 0.601   & \underline{80.31}   & 80.36 \\
\toprule
\multirow{2}{*}{\textbf{DIB (Ours)}}  & 0.421 & \textbf{0.625} & \textbf{81.44} & \textbf{81.63} \\
       & {\scriptsize ± 0.00} &{\scriptsize ± 0.01} & {\scriptsize ± 0.52} & {\scriptsize ± 0.53}\\
\bottomrule
\end{tabular}
\label{tab:SIMS performance}
\end{table}

\begin{table}[h]
\caption{Performance comparison on the MVSA-Single benchmark.}
\centering
\renewcommand{\arraystretch}{1}
\setlength{\tabcolsep}{6pt} 
\begin{tabular}{lcc}
\toprule
\textbf{Models} & \textbf{ACC-2 $\uparrow$} & \textbf{Weighted F1 $\uparrow$} \\
\midrule
HSAN$_{2017}$ ~\cite{xu2017analyzing}       & 69.88   & 66.90 \\
CoMN-Hop6$_{2018}$~\cite{xu2018co}     & 70.51   & 70.01 \\
MVAN-M$_{2020}$~\cite{yang2020image}       & 72.98   & 72.98 \\
MGNNS$_{2021}$~\cite{yang2021multimodal}     & 73.77   & 72.70 \\
CLMLF$_{2022}$~\cite{li2022clmlf}        & 75.33   & 73.46 \\
VSA-PF$_{2024}$~\cite{chen2024holistic}      & \underline{75.58}   & \underline{74.48} \\
CIGNN$_{2024}$~\cite{wang2024multimodal}       & 75.11   & 73.33 \\
\toprule
\multirow{1}{*}{\textbf{DIB (Ours)}} & \textbf{76.05} {\scriptsize ± 0.03} & \textbf{75.20} {\scriptsize ± 0.06} \\
\bottomrule
\end{tabular}
\label{tab:MVSA performance}
\end{table}

\section{Results and Analysis}
\subsection{Overal Performance}
We compare the proposed DIB with state-of-the-art baseline models on the MSA task in Table~\ref{tab:mosi&mosei}, ~\ref{tab:SIMS performance} and~\ref{tab:MVSA performance}, where the optimal results are highlighted in bold and the suboptimal results are marked with underlines. We also report the standard deviation of the reproduced models. The experimental results for CMU-MOSI dataset show that the proposed DIB outperforms advanced MSA models on the majority of the evaluation metrics, improving upon the competitive baseline MIB, which also leverages IB to learn representations, by 0.5\% in ACC-2, 0.5\% in F1-score, 1.3\% in MAE. Notably, these improvements are statistically significant, with p-values of $2.93 \times 10^{-4}$ (ACC-2), $9.29 \times 10^{-4}$ (F1), $8.5 \times 10^{-3}$ (MAE), and corresponding effect sizes (Cohen’s d) of 4.54, 4.21, and -1.40 respectively. Here, smaller p-values indicate stronger evidence against the null hypothesis, with values below 0.01 typically considered highly significant. Cohen’s d values further suggest substantial practical significance. Based on conventional benchmarks \cite{cohen1988}, d = 0.2 is regarded as a small effect, 0.5 as medium, and 0.8 as large. Therefore, the observed d values (greater than 4 or less than –1) reflect extremely strong differences between models. DIB also outperforms PS-Mixer by over 3\% in ACC-2. 

For CMU-MOSEI dataset, our model achieves comparable or superior prediction performance, particularly showing clear improvements over MIB, with gains of 0.6\% in ACC-7, 1.6\% in ACC-2, and 1.6\% in F1 score. The improvements are supported by strong statistical evidence, with p-values of $5.20 \times 10^{-7}$ (ACC-7), $1.50 \times 10^{-8}$ (ACC-2) and $1.93 \times 10^{-8}$ (F1), and corresponding effect sizes (Cohen’s d) of 7.367, 3.040, and 3.013, respectively, indicating large effects. In comparison with the best-performing model PS-Mixer, DIB achieves comparable overall performance and further yields improvements of 0.7\% in ACC-7 and 2.3\% in Corr, demonstrating its effectiveness in capturing multimodal sentiment signals.

Moreover, for CH-SIMS Chinese dataset, DIB obtains the best performance in terms of Corr, ACC-2, F1 evaluation metrics. Specifically, it outperforms the state-of-the-art baseline MUG remarkably by improving 2.4\% in Corr, 1.13\% in ACC-2, 1.27\% in F1-score, indicating DIB has the ability to generalize across different languages. When compared to another strong baseline EMT, the performance gains are statistically significant, with $p$-values of $0.004$ (Corr), $0.0002$ (ACC-2), and $5.88 \times 10^{-5}$ (F1), and corresponding effect sizes (Cohen’s $d$) of 2.71, 5.82, and 7.95, respectively. For MVSA-Single dataset, DIB continues to achieve the highest performance among all compared methods, with a 0.94\% improvement in ACC-2 and a 1.87\% improvement in Weighted F1 over the current baseline CIGNN, demonstrating its effectiveness on real-world visual-text sentiment tasks. To further validate these improvements, we conduct statistical tests against another open-source baseline CLMLF. The results show statistically significant gains, with $p$-values of $0.018$ (ACC-2) and $0.007$ (F1), and effect sizes (Cohen’s $d$) of 1.41 and 2.27, respectively, indicating moderate to large effects.

To sum up, these empirical observations demonstrate the effectiveness of DIB method on MSA tasks, implying the importance of learning comprehensive and effective representation. 

\begin{figure}
\centering 
\includegraphics[width=\linewidth]{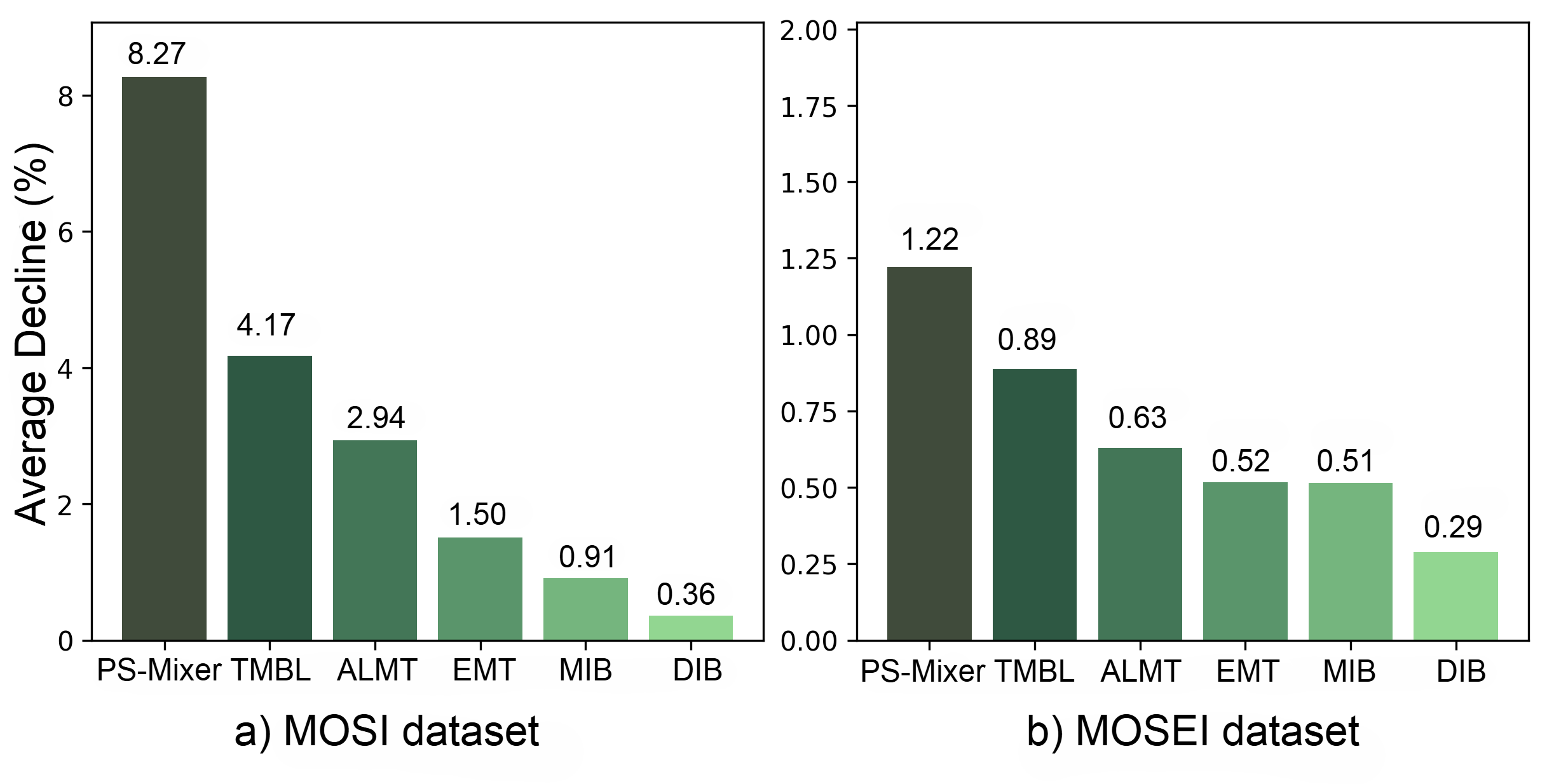} 
\caption{Average decline across five evaluation metrics in the noise addition experiment for the CMU-MOSI and CMU-MOSEI dataset. Similar trend is observed in CH-SIMS.}\label{fig:average decline}
\end{figure}

\begin{figure}
\centering 
\includegraphics[width=\linewidth]{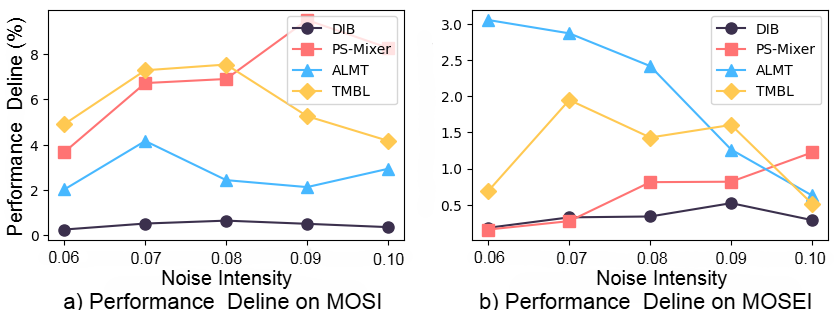} 
\caption{Comparison of average performance decline on the CMU-MOSI and CMU-MOSEI datasets under increasing multimodal noise intensities. For each level (0.06–0.10), 6–10\% of text tokens are randomly replaced or swapped, while Gaussian noise with corresponding standard deviations (0.06–0.10) is added to visual and acoustic features.}\label{fig:Noise Intensity}
\end{figure}

\subsection{Noisy Experiments}
\label{Sec:noisy experiments}
We evaluate the effect of noisy data on the prediction performance of MSA models. Following previous work~\cite{wei2019eda}, we generate noise data for text modality by applying random token replacement and position shuffling,  where 10\% of the tokens in each sequence are randomly perturbed on the training, validation, and test sets. For the audio and visual modalities, we sample Gaussian noise from the distribution $\mathcal{N}(0, 1)$ with a mean of zero and a variance of one to all data splits. Figure~\ref{fig:average decline} presents the degree of performance degradation of the learned models on noise data, measured by the average decline across five metrics (i.e. ACC-7, ACC-2, F1, Corr, MAE) on CMU-MOSI and CMU-MOSEI datasets.

As shown in Figure~\ref{fig:average decline}, our model has the lowest average decline of 0.36\% and 0.29\% across CMU-MOSI and CMU-MOSEI datasets, demonstrating the robustness and generalization ability of the proposed DIB under noisy environments. In comparison, the best-performing model PS-Mixer exhibits significantly higher average declines of 8.27\% and 1.22\% for CMU-MOSI and CMU-MOSEI. These promising results can be attributed to the robustness characteristics of DIB, which refines unimodal and multimodal representations by leveraging low rank entropy to filter out noise and redundancy, while avoiding the exchange of irrelevant information. In addition to methods that employ IB to learn compact representations (e.g. MIB), DIB still outperforms by achieving average decline that is 0.55\% and 0.22\% lower than MIB on CMU-MOSI and CMU-MOSEI dataset, respectively.
We also observe that the performance of the models on CMU-MOSEI dataset can achieve less decline than that of CMU-MOSI dataset. The possible reason behind this is that larger sample training sizes could facilitate model training to fit noise and reduce variance, thereby alleviating the negative impact of the noise and enhancing the robustness. Therefore, we suggest that a large number of samples is beneficial for achieving good generalization performance. Nevertheless, these experimental results validate the effectiveness and robustness of our method.

We further evaluate the prediction performance with various noise intensity on CMU-MOSI and CMU-MOSEI datasets by using average performance decline. From Figure~\ref{fig:Noise Intensity}, it is clear that the predictive performance of state-of-the-art models deteriorates under various noise intensity, while our method remaining relatively stable with a lower performance decline rate. Although the overall accuracy worsens, it is interesting to find that models (e.g. ALMT) experience a decline decrease as noise intensity increases. It is reasonable because noise in some cases acts as a form of data augmentation, potentially enhancing the model’s performance rather than degrading it. Rather than these models, DIB performs more consistently, further verifying the excellent robustness of our method to noisy data.\\

\begin{figure*}[!htbp]
\centering
\includegraphics[scale=0.65]{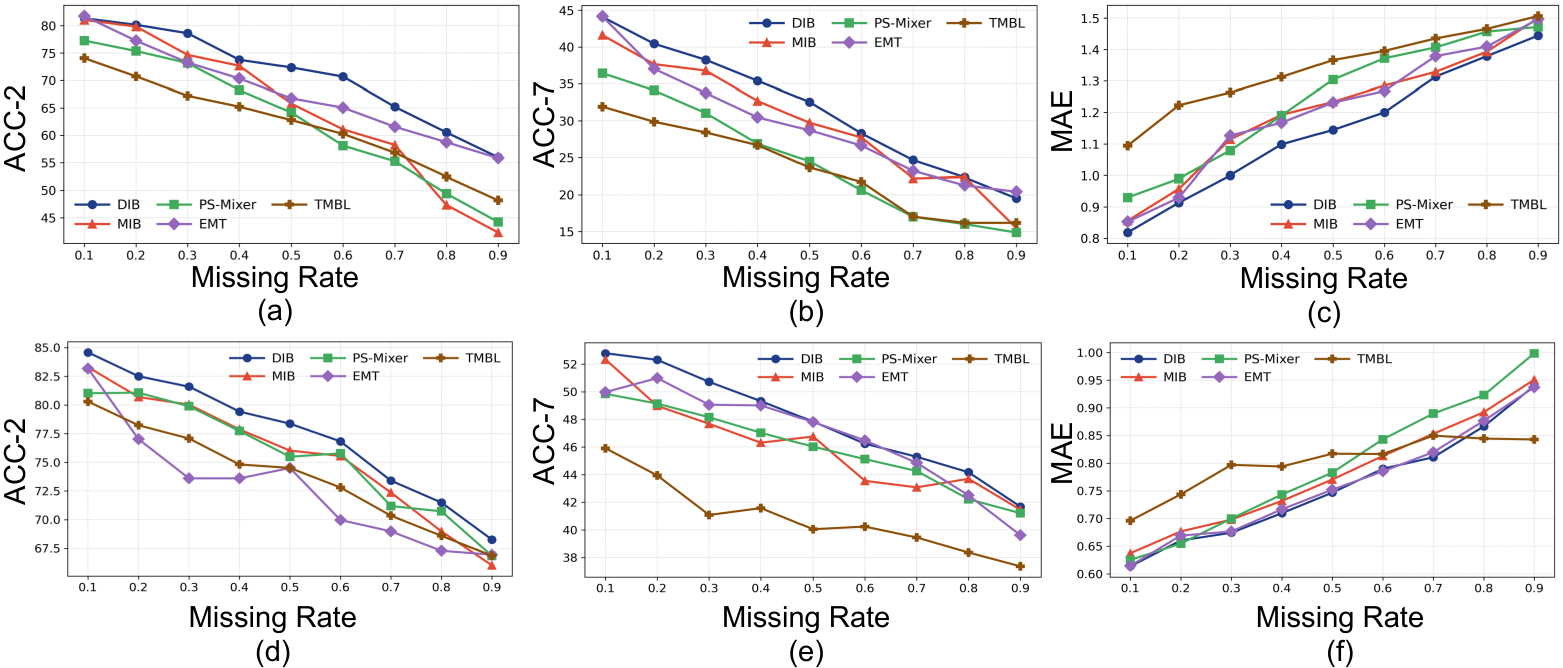}
\caption{Performance curves of various missing rates. (a), (b) and (c) are the ACC-2, ACC-7, MAE curves on MOSI dataset. (d), (e) and (f) are the ACC-2, ACC-7, MAE curves on MOSEI. Note: The smaller MAE indicates the better performance.}
\label{fig:missing_line}
\end{figure*}

\subsection{Missing Experiments}
As illustrated in Figure~\ref{fig:missing_line}, we evaluate the robustness of different models under varying missing modality rates on both CMU-MOSI and CMU-MOSEI datasets. Specifically, we simulate increasing percentages of missing modalities by randomly masking during training, validation and testing, and track performance trends across key metrics: ACC-2, ACC-7, and MAE. The missing rates range from 0.1 to 0.9 with an interval of 0.1, meaning that at most 90\% of modality information may be missing in the most extreme cases. Subfigures (a)–(c) show the results on MOSI, while (d)–(f) correspond to MOSEI. The curves clearly indicate that our proposed DIB method consistently achieves strong performance across all missing rates, particularly when the missing ratio becomes severe.  Note that lower MAE and higher ACC scores indicate better performance, and DIB achieves competitive results in both aspects, validating its robustness in real-world multimodal degradation scenarios.

\subsection{Efficiency Analysis}
We further evaluate the computational efficiency of our proposed DIB model and compare it with representative baselines on the CMU-MOSI dataset, as summarized in Table~\ref{tab:efficiency_mosi}. 
All models are trained under the same experimental setup and batch size of $32$. 
We report three key indicators: the number of trainable parameters (in millions), the average wall-clock training time per epoch (in seconds), and the peak GPU memory usage (in MiB). As shown in Table~\ref{tab:efficiency_mosi}, DIB achieves a comparable computational footprint to baselines while offering substantially better robustness and comparative performance. Although MIB with 109.8M parameters shows shorter per-epoch time of $3.13$\,s due to its simple feature concatenation strategy, DIB incorporates a more expressive bottleneck attention fusion mechanism with only a slight increase in parameter scale. Compared with larger or more complex backbones such as TMBL and EMT, DIB demonstrates lower or comparable computational cost with around 27.7\% faster training than TMBL and 15.3\% faster than EMT, while also reducing peak memory usage by 32.5\% and 3.1\%, respectively.


\begin{table}[htbp]
\centering
\renewcommand{\arraystretch}{1}
\setlength{\tabcolsep}{1pt}
\caption{Efficiency comparison of different models on the CMU-MOSI dataset under the same batch size. Metrics include number of parameters (in millions), average training time per epoch, and peak GPU memory usage.}
\begin{tabular}{lccc}
\toprule
\textbf{Model} & \textbf{\#Params (M)} & \textbf{Time/Epoch (s)} & \textbf{Memory (MiB)} \\
\midrule
MIB       & 109.8  & 3.13   & 4920 \\
PS-Mixer  & 110.2  & 13.34  & 4926 \\
EMT       & 103.8  & 15.49  & 5220 \\
TMBL      & 319.5  & 18.15  & 7498 \\
DIB (Ours) & 109.8  & 13.12  & 5058 \\
\bottomrule
\end{tabular}
\label{tab:efficiency_mosi}
\end{table}


\begin{table}[h]
\caption{Ablation study results on CMU-MOSI and CMU-MOSEI datasets. (-) represents removal for the mentioned factors. Model 1,2 present the effect of essential component; Model 3,4,5,6,7 present the effect of LRIB; model 8,9,10 exhibit the effect of modalities; model 11,12,13 depict the effect of dominant modality.}\label{ablation study}
\setlength{\tabcolsep}{3.2pt} 
\renewcommand{\arraystretch}{1} 
\centering
\begin{tabular}{lcccc}
\hline
\small
\multirow{2}{*}{Model} & \multicolumn{2}{c}{MOSI} & \multicolumn{2}{c}{MOSEI} \\ \cline{2-5} 
            & MAE (↓)   & F1 (↑)     & MAE (↓)   & F1 (↑)     \\ \hline
\rowcolor{gray!20}
\textbf{ Ours}        & \textbf{0.715} & \textbf{85.56} & \textbf{0.588} & \textbf{85.97} \\ \hline
1) (-) LRIB module     & 0.739    & 83.53 & 0.590    & 84.96 \\
2) (-) Fusion module     & 0.851    & 81.40 & 0.624    & 82.97 \\ \hline
3) (-) LRIB on Text     & 0.738    & 83.93 & 0.595    & 85.08 \\
4) (-) LRIB on Audio     & 0.736    & 84.10 & 0.596    & 84.63 \\
5) (-) LRIB on Visual     & 0.736    & 83.67 & 0.595    & 84.70 \\
6) (-) Unimodal LRIB     & 0.749    & 83.40 & 0.595    & 85.47 \\
7) (-) Multimodal LRIB     & 0.734    & 84.03 & 0.594    & 85.15 \\ \hline
8) (-) Text $t$  & 1.503    & 50.57 & 0.982    & 59.09 \\
9) (-) Visual $v$   & 0.734    & 83.26 & 0.599    & 84.06 \\
10) (-) Audio $a$    & 0.746    & 82.92 & 0.595    & 85.13 \\ \hline
11) Audio-dominant    & 0.747    & 83.44 & 0.601    & 84.78 \\
12) Visual-dominant     & 0.734    & 83.35 & 0.599    & 83.72 \\ 
13) All Modalities    & 0.742    & 83.75 & 0.592    & 84.87 \\ \hline
\end{tabular}\label{tab:ablation_study}
\end{table}

\subsection{Ablation Studies}
\label{sec:ablation}
In this section, we conduct comprehensive ablation studies and comparison experiments to assess the effectiveness of the components in DIB.

\noindent{\textbf{Effect of Essential Component.} We conduct ablation studies to evaluate the effect of removing core modules, with results shown in  Table~\ref{tab:ablation_study} (1)-(2). Specifically, Table~\ref{tab:ablation_study} (1) removes the LRIB objective entirely, while Table~\ref{tab:ablation_study} (2) removes the fusion module and replaces it with simple feature concatenation. Both degradations lead to a notable drop in performance across MOSI and MOSEI datasets, especially the fusion module, which results in a large increase in MAE and a substantial decrease in F1 score. This result highlights the critical role of both the LRIB objective and our designed bottleneck fusion strategy in achieving strong performance.}

\noindent{\textbf{Effect of LRIB.} We investigate the contribution of LRIB by selectively removing it from different parts of the architecture. Table~\ref{tab:ablation_study} (3)–(5) denote the exclusion of LRIB from individual modalities, while Method (6) removes all unimodal LRIB objectives, and Method (7) removes only the multimodal LRIB. The results demonstrate that each modality benefits from the guidance of LRIB, with text showing the most significant contribution. Moreover, combining both unimodal and multimodal LRIB objectives leads to the best performance, underscoring the necessity of using LRIB at both levels to effectively extract and refine modality-specific and fused representations.}

\noindent{\textbf{Effect of Modalities.} We study the impact of individual modality on the generalization performance. We present the prediction performance of multimodal learning without a certain modality illustrated in Table~\ref{tab:ablation_study} (8)-(10). It is obvious that DIB without text modality has the worst performance compared with other modalities. This implies that textual content can provide richer and more accurate semantic information to improve generalization performance for MSA tasks. At the same time, the performance drop caused by removing audio or visual modalities suggests that they provide complementary information that benefits overall prediction.}

\noindent{\textbf{Effect of Dominant Modality.} As previously elucidated, text modality plays a crucial role in MSA tasks. To further explore its efficacy, we conduct experiments with three alternative modality configurations: audio-dominant, visual-dominant, and non-dominant (denoted as "All Modalities" in Table~\ref{tab:ablation_study}). The empirical observations in Table~\ref{tab:ablation_study} (11)-(13) suggest that employing either audio or visual modalities as the primary guide, or utilizing a non-dominant approach, results in suboptimal performance compared to the text-dominant strategy implemented in the proposed method.}

\begin{table}[h]
\caption{Comparison results of fusion methods on CMU-MOSI and CMU-MOSEI datasets. We also present the average time per epoch for the comparative attention fusion methods and DIB.}\label{ablation study}
\setlength{\tabcolsep}{3.2pt} 
\renewcommand{\arraystretch}{1} 
\centering
\begin{tabular}{lcccc}
\hline
\small
\multirow{2}{*}{Model} & \multicolumn{2}{c}{MOSI} & \multicolumn{2}{c}{MOSEI} \\ \cline{2-5} 
            & MAE (↓)   & F1 (↑)     & MAE (↓)   & F1 (↑)     \\ \hline
            \rowcolor[HTML]{FBF4DB}
            \multicolumn{5}{l}{Simple Manipulation} \\ \hline
1) Concatenation   & 0.851    & 81.40 & 0.624    & 82.97 \\ 
2) Addition   & 0.867    & 80.37 & 0.611    & 83.47 \\ \hline
            \rowcolor[HTML]{FBF4DB}
            \multicolumn{5}{l}{Tensor Fusion} \\ \hline
3) TFN~\cite{zadeh2017tensor}   & 0.893    & 80.53 & 0.607    & 83.29 \\ 
4) LMF~\cite{liu2018efficient}   & 0.734    & 84.36 & 0.613    & 84.95 \\ \hline
            \rowcolor[HTML]{FBF4DB}
            \multicolumn{5}{l}{Graph-based Fusion} \\ \hline
5) ARGF~\cite{mai2020modality}    & 0.751    & 84.31 & 0.601    & 84.27 \\
6) GraphCAGE~\cite{wu2021graph}    & 0.778    & 84.48 & 0.622    & 85.13 \\ \hline
            \rowcolor[HTML]{FBF4DB}
            \multicolumn{5}{l}{Attention Fusion} \\ \hline
7) MSAF~\cite{su2020msaf}  & 0.817    & 84.14 & 0.605    & 85.20 \\
\,\,\,\,\,\,\,\scriptsize{\textit{Time}} &\multicolumn{2}{r}{\scriptsize \textit{14.59 s}} & \multicolumn{2}{r}{\scriptsize \textit{683.570 s}} \\
8) MMMU-BA~\cite{ghosal2018contextual}   & 0.787    & 84.45 & 0.599    & 84.77 \\ 
\,\,\,\,\,\,\,\scriptsize{\textit{Time}} &\multicolumn{2}{r}{\scriptsize \textit{16.41 s}} & \multicolumn{2}{r}{\scriptsize \textit{641.493 s}} \\\hline
\rowcolor{gray!20}
\textbf{ Ours}        & \textbf{0.715} & \textbf{85.56} & \textbf{0.588} & \textbf{85.97} \\ 
\,\,\,\,\,\,\,\scriptsize{\textit{Time}} &\multicolumn{2}{r}{\scriptsize \textit{13.05 s}} & \multicolumn{2}{r}{\scriptsize \textit{639.643 s}} \\
$\Delta SOTA$ &$\uparrow$0.019 &$\uparrow$1.08 &$\uparrow$0.011 &$\uparrow$0.77 \\ \hline
\end{tabular}\label{tab:fusion comparison}
\end{table}

\subsection{Fusion Techniques Comparison}
We conduct extensive comparative experiments to investigate the role of several mainstream fusion approaches in the generalization ability encompassing four representative fusion methods: simple manipulation, tensor fusion, graph-based fusion and attention fusion. More specifically, concatenation and addition are among the simplest methods for directly combing multimodal features linearly without explicit interaction modeling. In contrast, tensor fusion~\cite{zadeh2017tensor,liu2018efficient} converts the initial embedding into high-dimensional tensors and then compute the tensor product to capture higher-order interactions across modalities. Graph-based fusion~\cite{mai2020modality,wu2021graph} organizes multimodal information into a graph structure to aggregate and propagate information more interpretably and effectively, while the prevailing attention fusion~\cite{su2020msaf,ghosal2018contextual} utilizing attention mechanisms to weigh the importance of different interaction parts. As shown in Table~\ref{tab:fusion comparison}, replacing the proposed attention bottleneck fusion with other fusion methods results in a performance decline across all evaluation metics, which highlights the effectiveness of leveraging multimodal complementary information and the excellent generalization ability of our proposed bottleneck fusion module. In particular, for the F1-score metric, DIB achieves a 1.08\% improvement on the CMU-MOSI dataset and a 0.77\% improvement on the CMU-MOSEI dataset. Furthermore, we compare comparative models (7) and (8) in Table~\ref{tab:fusion comparison}, which also utilize attention mechanisms, with our proposed model in terms of computational efficiency. Notably, DIB demonstrates lower time consumption per epoch, requiring 13.05 seconds on CMU-MOSI and 639.64 seconds on CMU-MOSEI dataset. In contrast, MSAF~\cite{su2020msaf} takes 14.59 seconds on CMU-MOSI and 683.57 seconds on CMU-MOSEI dataset every epoch. This time reduction indicates that the introduction of bottleneck embeddings effectively alleviates the attention computation overhead while achieving superior results by filtering out superfluous information.

\begin{figure}
\centering 
\includegraphics[width=\linewidth]{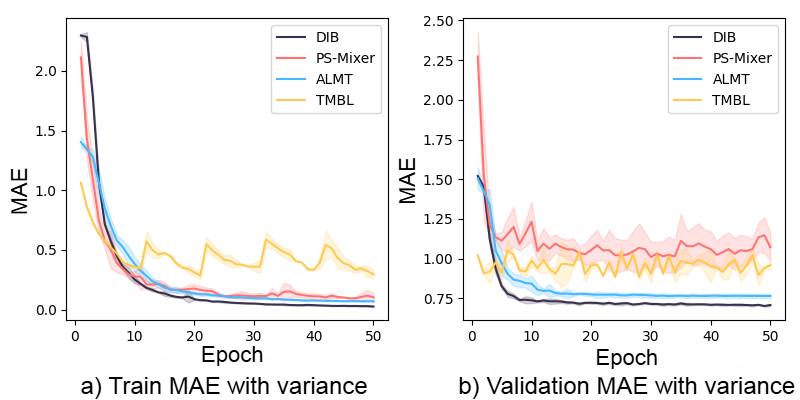} 
\caption{Comparison of convergence performance on the training and validation sets of the CMU-MOSI dataset. The shaded areas represent the variance in results obtained from multiple random seeds. }\label{convergence performance}
\end{figure}

\begin{figure}
  \centering
  \includegraphics[width=0.98\linewidth]{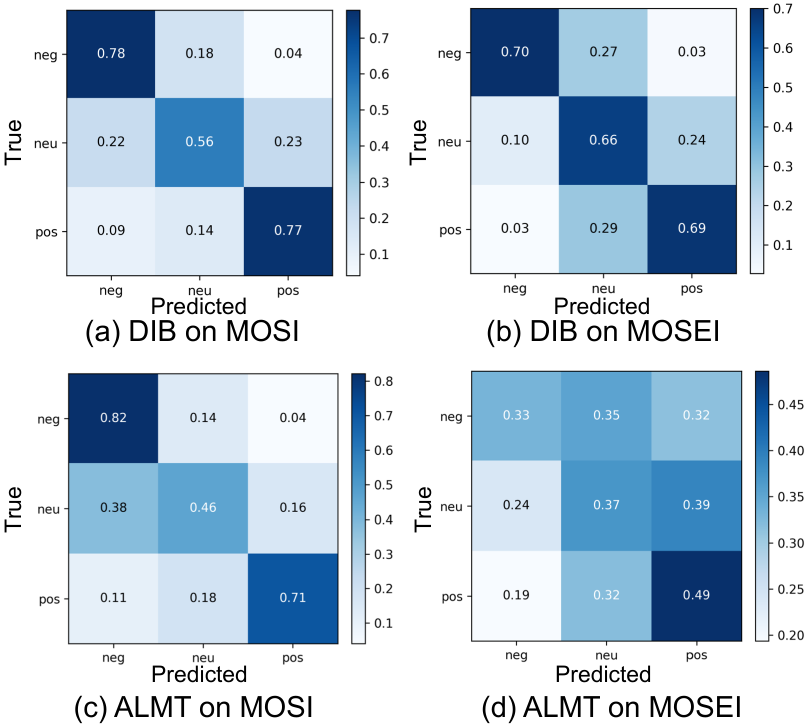}
  \caption{Three-class confusion matrices, i.e., positive, neutral, negative, of DIB and ALMT on the MOSI and MOSEI datasets.}
  \label{fig:confusion_matrix}
\end{figure}

\begin{figure}
  \centering
  \includegraphics[width=0.89\linewidth]{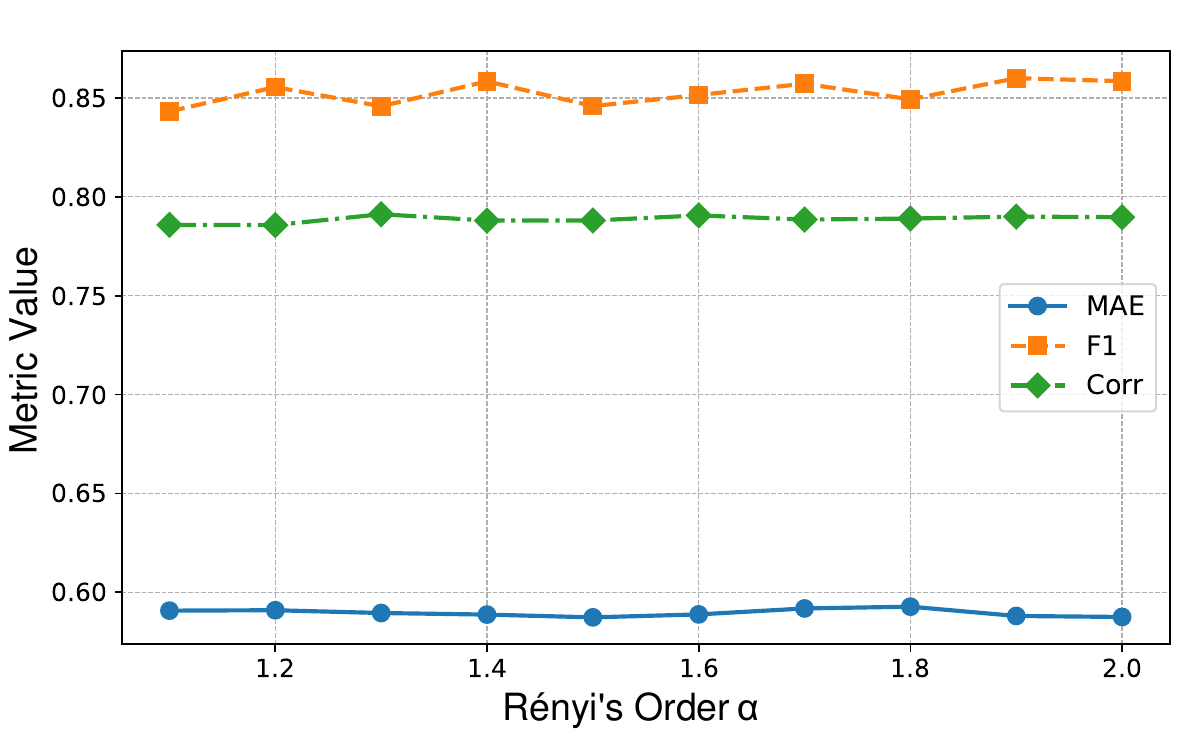}
  \caption{Sensitivity analysis of the R\'enyi’s order $\alpha$ on the MOSI and MOSEI datasets. The performance remains stable across a wide range of $\alpha$ values, validating the robustness of our entropy surrogate.}
  \label{fig:para_sensitivity}
\end{figure}

\subsection{Further Analysis}
\noindent{\textbf{Convergence Performance Comparison.}} We compare the convergence speed of DIB and state-of-the-art models on the CMU-MOSI dataset in Figure~\ref{convergence performance}. Similar trends can be observed in CMU-MOSEI and CH-SIMS dataset. While DIB, PS-Mixer, and ALMT have the similar convergence speed during training, DIB achieves the lowest MAE as shown in Figure~\ref{convergence performance} (a). For the verification process, Figure~\ref{convergence performance} (b) shows that DIB not only exhibits the fastest convergence speed but also achieves the best prediction performance. In addition, the experimental results demonstrate the stability of the proposed DIB, as reflected by its lower variance depicted in the shaded area. All of the above experimental observations further verify the computational efficiency and robustness of our method.

\noindent{\textbf{Neutral Class Analysis.} To more thoroughly assess model behavior across sentiment categories, we include three-way confusion matrices (positive, neutral, negative) for both the MOSI and MOSEI datasets, as shown in Figure~\ref{fig:confusion_matrix}. Compared to the latest baseline model ALMT, our proposed DIB demonstrates improved classification balance, especially in identifying neutral sentiments. This result highlights DIB’s ability to capture subtle or less polar cues across modalities, contributing to more interpretable sentiment predictions.}

\noindent\textbf{Sensitivity Analysis on R\'enyi’s Order $\alpha$.} We further conduct a sensitivity analysis on the R\'enyi’s order $\alpha$ to examine its influence on model performance. Following the setup in Section~\ref{sec:Implementation Details}, we vary $\alpha$ from $1.1$ to $2.0$ in increments of $0.1$, while keeping all other hyperparameters fixed. As illustrated in Figure~\ref{fig:para_sensitivity}, the overall performance of DIB remains stable across a broad range of $\alpha$ values, indicating that the model is not overly sensitive to this parameter. Notably, $\alpha=1.9$ consistently yields the best trade-off between stability and predictive accuracy, and is thus adopted as the default configuration in all reported experiments.

\begin{figure*}[!htb]
\centering
\includegraphics[scale=0.55]{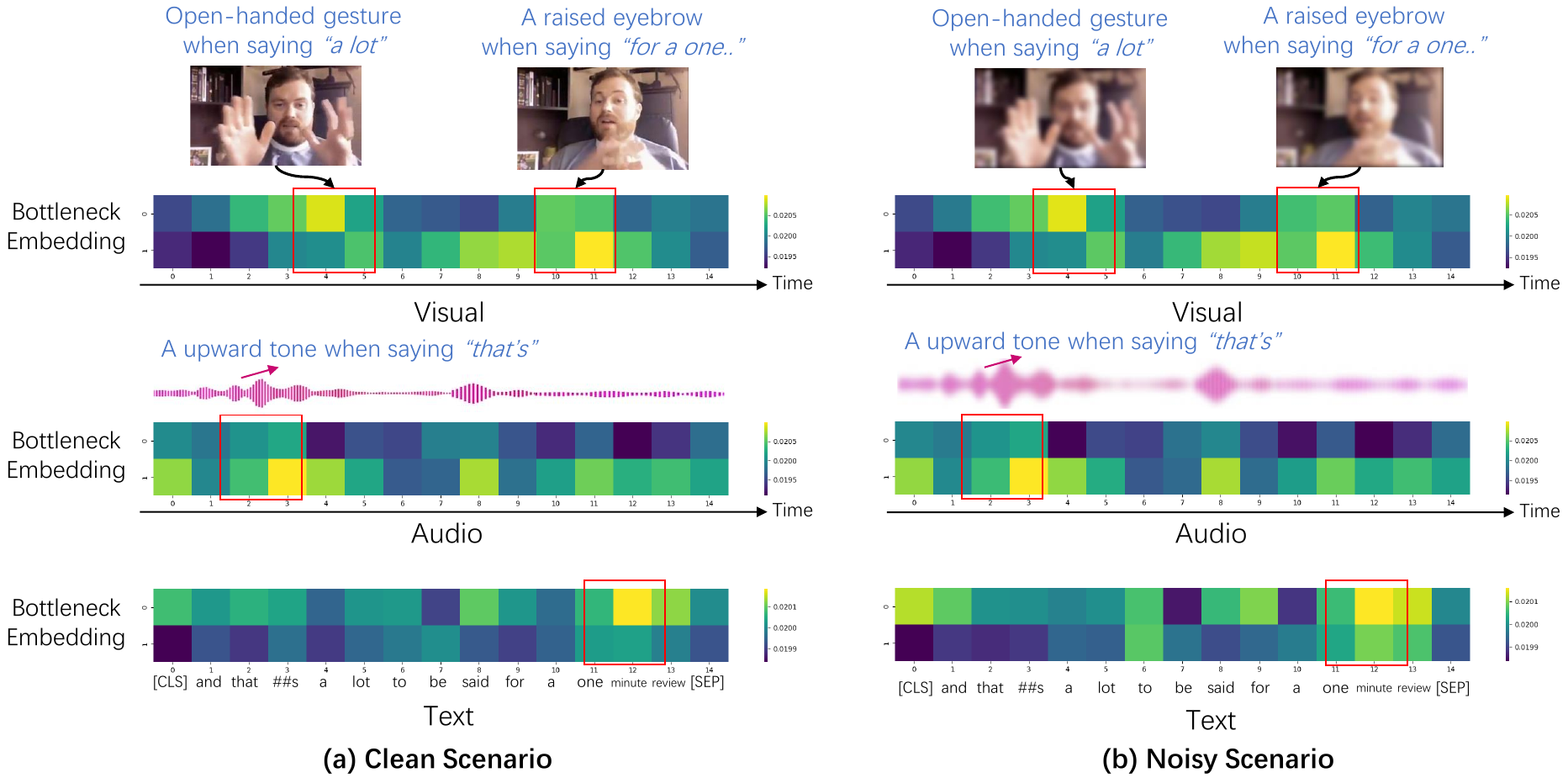}
\caption{Visualization of attention heatmap where the bottleneck embeddings is the target and each unimodal feature is the source in the clean and noisy scenario. The lighter the color, the higher the attention score. Key regions with the strongest attention are further highlighted with red rectangles. In both scenarios, we find that the bottleneck embeddings effectively learns to focus on essential parts of the sentence that contribute to the sentiment (e.g. a raised eyebrow, gesture and the lifted tone).}
\label{fig:attention}
\end{figure*}

\begin{figure}
\centering 
\includegraphics[width=\linewidth]{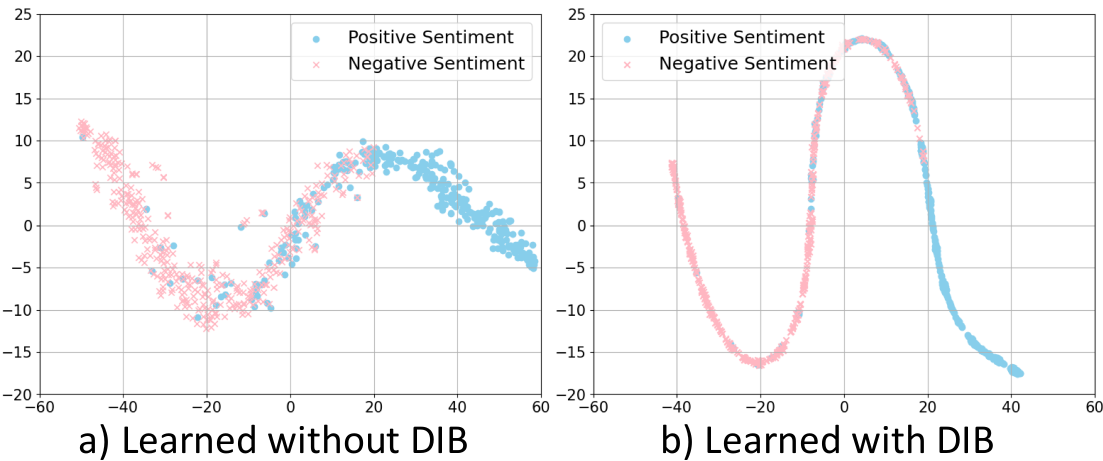} 
\caption{Visualization of multimodal representations without and with DIB for CMU-MOSI dataset. Positive samples are marked as blue dots, and negative samples as pink crosses.}\label{tsne}
\end{figure}

\subsection{Visualization}
\noindent{\textbf{Visualization of Attention Heatmap.} we empirically investigate the signals DIB captures by visualizing the attention weights of bottleneck embeddings across each unimodal representation under both clean and noisy settings, as shown in Figure~\ref{fig:attention}. We set the length of the bottleneck embeddings as 2 and the sample is selected from the test set of CMU-MOSI. For clearer interpretation, we present key video frames, audio waveforms, text tokens from BERT, and alignment information. The three modalities (visual, audio, text) are aligned by words, therefore the horizontal axes of the plots are of equal length. From the figure, we can observe that DIB captures meaningful interactions. Specifically, in a clean condition (Figure~\ref{fig:attention} (a)), when considering visual modality, the bottleneck embeddings pay the most attention to those visual cues where the speaker makes an open-handed gesture and raises the eyebrow, indicating emphasis and positive attitude. In the audio modality, DIB attends to the upward intonation conveying affirmation. For the text, the attention scores for the phrase "one minute review" are higher, likely due to its importance in shaping the overall meaning of the sentence. Although the text alone may appear neutral, the accompanying visual and acoustic cues reinforce the positive sentiment, demonstrating DIB’s ability to leverage cross-modal complementarity to resolve textual ambiguity. And DIB successfully perceives these implicit cues for performance improvement. In terms of the noisy setting (Figure~\ref{fig:attention} (b)), after introducing Gaussian noise to the unimodal representation, DIB still identifies these essential signals, successfully classifying the sentiment as positive, which once again verifies that DIB is robust against noise in the data.

\noindent{\textbf{Visualization of Multimodal Representations.} We provide visualization for feature distributions of the multimodal representations to intuitively observe the robustness and effectiveness of the proposed DIB. Following previous work~\cite{van2008visualizing}, we apply the t-SNE algorithm to visualize the feature distribution, which projects the high-dimensional representation into a 2-dimensional feature space to capture the local structure of high-dimensional data. Figure~\ref{tsne} presents the embedding space learned without or with DIB. Positive samples are marked as blue dots, and negative samples as pink crosses.} From Figure~\ref{tsne}, we observe that the multimodal representations in the embedding space learned without DIB are more scattered, leading to inefficiency in forming distinguishing clusters for each class. In contrast, the representations from the same class learned using DIB can form discriminative clusters. The visualization results reveal that the proposed DIB method can be immune to noise interference and sufficiently leverage discrepancy information of unimodality to capture discriminative representation across multiple modalities, thereby enhancing robustness and generalization ability.

\begin{figure}
\centering 
\includegraphics[width=\linewidth]{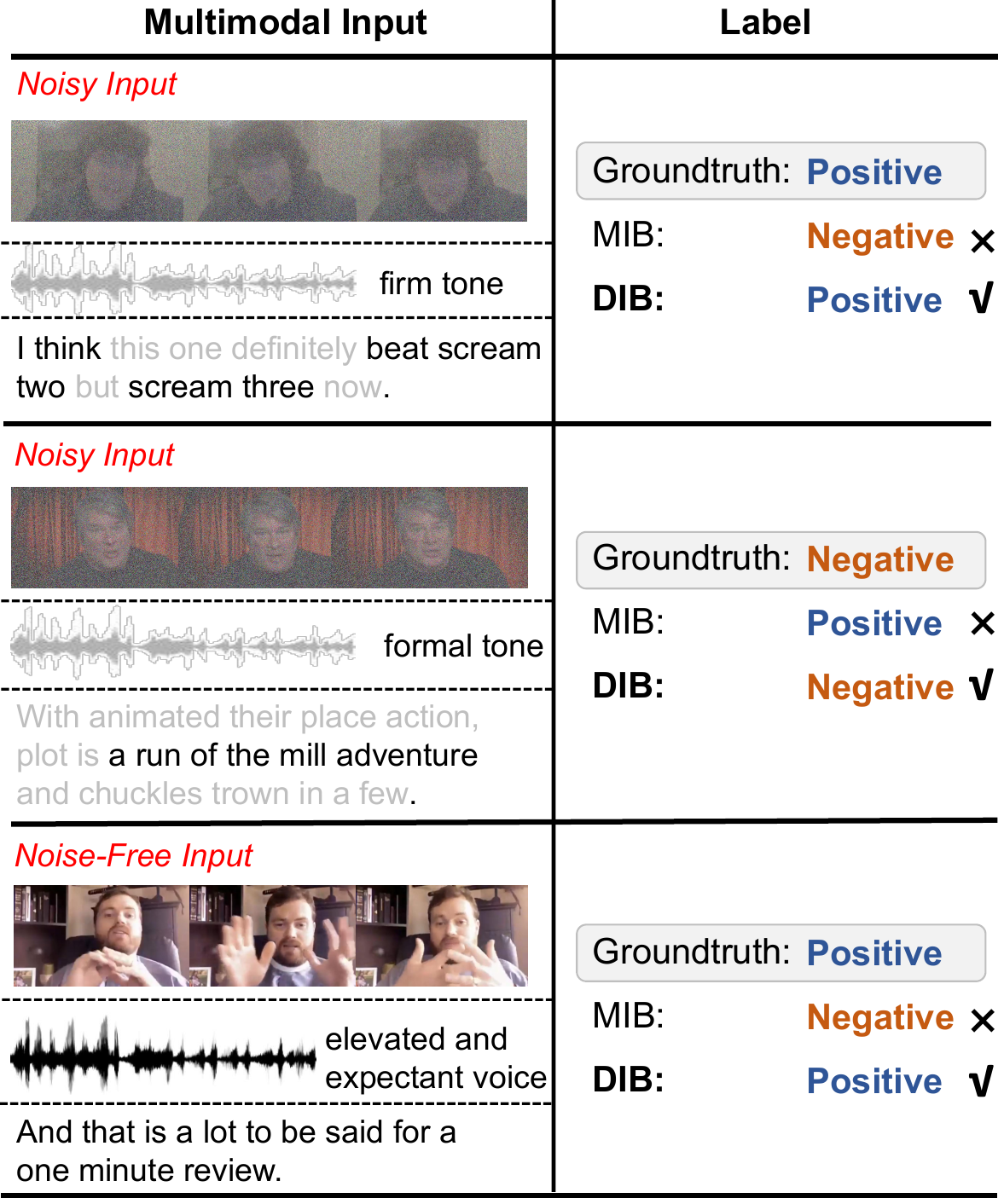} 
\caption{Real examples including noisy input and noise-free input case from the CMU-MOSI dataset. For each example, we present the ground truth label and prediction output of MIB~\cite{mai2022multimodal} and our proposed DIB model. Note: The noisy input (i.e. text scripts with errors) is visualized to facilitate readers' understanding. Random and various noise settings are applied to the original input sequence, as described in Section~\ref{Sec:noisy experiments}.}\label{fig:case study}
\end{figure}

\begin{figure*}
\centering
\includegraphics[scale=0.48]{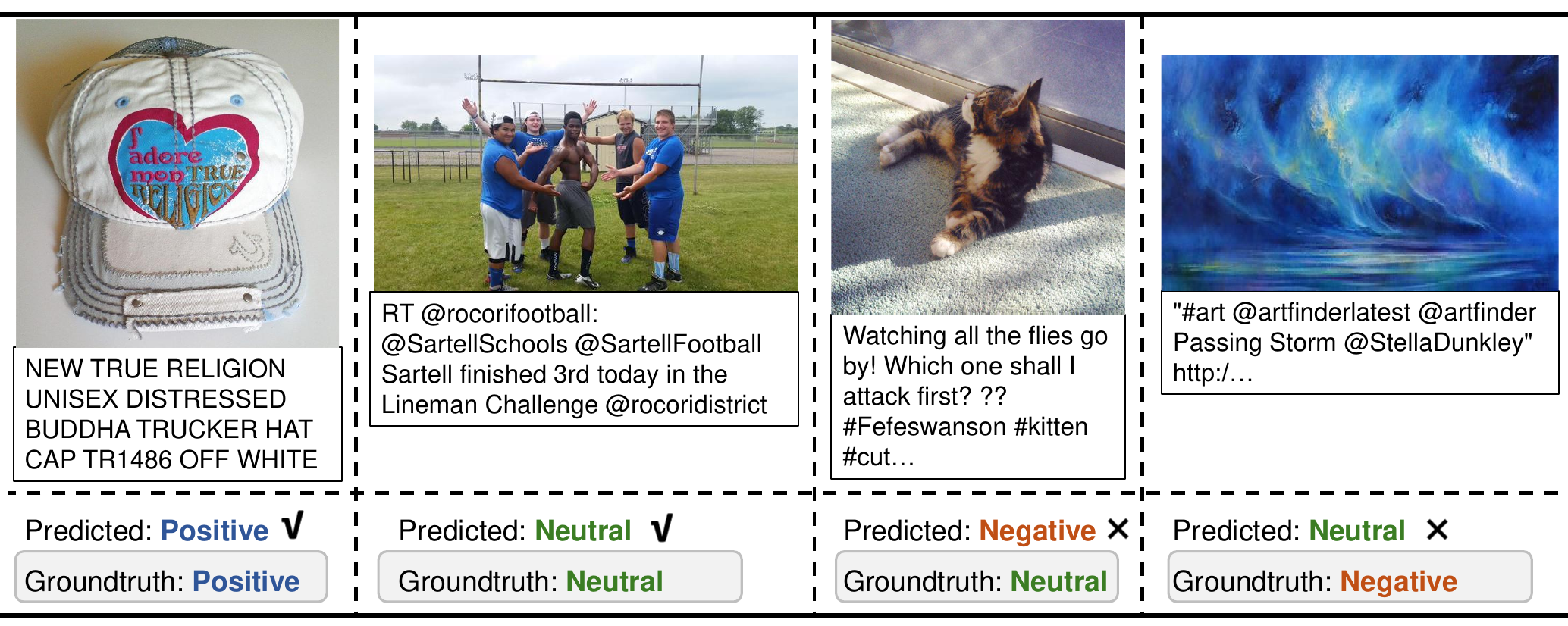}
\caption{Case Investigation including failure cases on the MVSA-Single dataset. The left two examples are correctly predicted as positive and neutral, respectively. The right two examples illustrate failure cases where sentiment was misclassified due to misleading lexical cues and challenges in interpreting abstract visual content.}
\label{fig:MVSA_cases}
\end{figure*}

\subsection{Case Study}
To demonstrate the efficacy of DIB, we show real examples in both noisy and clean scenarios made by MIB~\cite{mai2022multimodal} and DIB from the CMU-MOSI dataset, as shown in Figure~\ref{fig:case study}. DIB successfully identifies emotions in these complicated and confusing samples where MIB struggles due to DIB's ability to preserve salient information while ignoring non-essential details. On one hand, DIB exhibits robustness against noise, as evidenced by the first and second samples. In the first case, although the word "beat" in the text indicates a negative sentiment, DIB is still able to perceive sentiment cues and make the correct positive prediction. In the second case, the speaker has no distinct facial expression to suggest his emotional inclination, and his tone remains neutral and unexpressive. However, DIB effectively infers sentiment cues from the language modality by interpreting the phase "a run of the mill". On the other hand, DIB can incorporate complementary information across multiple modalities to enhance performance. For instance, in the third example, where language is ambiguous and lacks clear emotional indicators, DIB successfully leverages visual cues (e.g. a raised eyebrow and gesture) and acoustic signals (e.g. a lifted and expectant voice) to accurately predict sentiment, showcasing its ability to integrate discriminative features from both textual and non-textual modalities.

\section{Discussion}
In our work, we adopt low-rank R\'enyi entropy over Shannon entropy. We provide comprehensive justification for this choice. (1) Direct estimation capability: It can be computed directly from data samples without requiring knowledge of the underlying probability density function, which is particularly valuable when working with complex, high-dimensional datasets. Additionally, fast approximations like Random projection and Input-Sparsity Transform, and Sparse Graph Sketching are also developed. In contrast, Shannon entropy depends on estimating the full distribution, which is often unreliable in practice. (2) Mathematical completeness: The corresponding union entropy, conditional entropy, and mutual information are all well-defined within this framework, providing a comprehensive information-theoretic toolkit. (3) Enhanced robustness: As demonstrated in Theorem 1 of \cite{dong2023robust}, eliminating relatively small eigenvalues enhances the robustness of the entropy measure against noise and perturbations, which is crucial for reliable information quantification in real-world scenarios. In contrast, Shannon entropy or other traditional entropy estimations does not inherently offer such robustness, as it treats all components equally regardless of their contribution or stability.

We provide the analysis of comparison between our bottleneck fusion mechanism and several representative fusion strategies, including simple element-wise operations, high-dimensional tensor fusion, graph-based cross-modal modeling, and attention-based fusion. Our method consistently outperforms these baselines due to its ability to balance informativeness and compactness. Simple operations such as concatenation or addition treat all modality features equally, making them prone to noise and modality-specific biases. In contrast, tensor fusion increases expressive power by modeling complex interactions, but it introduces substantial computational overhead and risks of overfitting, particularly in low-resource scenarios. Graph-based methods rely on explicitly defined graph structures, which may struggle to capture dynamic or implicit relationships between modalities. Attention-based fusion improves flexibility in modeling interactions but often lacks mechanisms to suppress redundancy, potentially retaining irrelevant or noisy features. Our proposed bottleneck fusion addresses these limitations by imposing a capacity constraint that selectively allows only the most informative features to pass through. This not only enhances robustness against noise and redundancy but also reduces computational overhead, leading to more efficient and effective cross-modal representation learning.

The observed improvements in accuracy and robustness suggest that DIB holds strong potential for real-world applications such as video social media analysis and interactive systems, where noisy and unpredictable inputs are common. To further understand the model's behavior under real-world conditions, we also investigate several representative cases from the MVSA-Single dataset, which reflect the characteristics of online environments. As illustrated in Figure \ref{fig:MVSA_cases}, the left two samples are correctly classified while the other two are misclassified. Despite the inherent label imbalance of the dataset, with the neutral category being underrepresented, the second example shows that our model is still able to robustly predict this class, suggesting its ability to learn more accurate and expressive representations even under limited data conditions. In the failure cases, the third sample is misclassified as non-neutral, likely due to the presence of sentiment-laden words such as “attack,” which the model may overemphasize in the absence of sufficient visual context. In the fourth sample, the model struggles to interpret abstract or artistic visuals. For example, in the case of a stormy figure that conveys a turbulent atmosphere through dark color tones, it incorrectly predicts the emotion as neutral, whereas the correct label is negative. These limitations point to potential areas for improvement, which are further discussed in the following section.

\section{Conclusion and Future Work}
In this paper, we present the Double Information Bottleneck (DIB) framework, a robust and efficient model for MSA. By integrating low-rank R\'enyi's entropy, DIB effectively tackles the challenges of noise and redundancy in multimodal data. Unlike the traditional entropy measure, our low-rank approach offers computational tractability and enhanced robustness by focusing on the most informative eigenvalues. Moreover, we innovatively design the attention bottleneck fusion to achieve superior results while enhancing computational efficiency by preventing the exchange of noise and redundant information. The framework's dual focus on learning compact and informative unimodal representations and preserving critical cross-modal correlations enables the construction of a resilient, unified multimodal representation. Extensive empirical validation, including quantitative results and visualization, confirms that DIB not only outperforms state-of-the-art methods in prediction accuracy, but also demonstrates exceptional robustness across a variety of challenging conditions, such as additive noise, cross-modal misalignment, and missing modality. These experiments suggest that DIB is well-suited for practical multimodal systems like sentiment-aware recommendation systems and multimodal conversational agents, where incomplete or noisy signals frequently arise.

In future work, we intend to address several specific limitations and explore concrete directions to refine and generalize our approach. First, when using global multimodal labels to supervise unimodal representation learning, the multimodal label may hider unimodal representations from extracting more discriminative and precise information, as discussed in~\cite{mai2024meta}. To mitigate this issue, a promising direction is to incorporate automatic or adaptive label learning techniques, such as modality-specific pseudo-label generation or meta-learning-based supervision strategies, which thereby enhances the learning quality of the LRIB module. Second, our analysis of failure cases suggests that the model sometimes over-relies on sentiment-heavy lexical tokens without sufficient grounding in corresponding visual evidence, especially when dealing with abstract, artistic, or subtle visual content. To address this, we plan to explore visual grounding or vision-language alignment techniques. By linking sentiment expressions in text (e.g., "thrilling") to concrete visual cues (e.g., facial expressions), the model can better learn to validate or refute sentiment predictions based on grounded visual support. Lastly, given the modularity of our representation learning framework, we hope to explore its applicability to broader multimodal tasks beyond sentiment analysis, including Visual Question Answering and Text-to-Video Retrieval, where robust and discriminative multimodal representations are equally critical.


\section*{Acknowledgments}
This work has been supported by the National Research Foundation Singapore under AI Singapore Programme (Award Number: AISG-GC-2019-001-2A and AISG2-TC-2022-004)


\bibliographystyle{unsrt}

\bibliography{ref.bib}

\begin{thebibliography}{10}

\bibitem{camben}
Erik Cambria, Devamanyu Hazarika, Soujanya Poria, Amir Hussain, and RBV Subramanyam.
\newblock Benchmarking multimodal sentiment analysis.
\newblock In {\em {CICLing}}, pages 166--179, 2017.

\bibitem{rideaux2021multisensory}
Reuben Rideaux, Katherine~R Storrs, Guido Maiello, and Andrew~E Welchman.
\newblock How multisensory neurons solve causal inference.
\newblock {\em Proceedings of the National Academy of Sciences}, 118(32):e2106235118, 2021.

\bibitem{mao2022biases}
Rui Mao, Qian Liu, Kai He, Wei Li, and Erik Cambria.
\newblock The biases of pre-trained language models: An empirical study on prompt-based sentiment analysis and emotion detection.
\newblock {\em IEEE transactions on affective computing}, 14(3):1743--1753, 2022.

\bibitem{he2022meta}
Kai He, Rui Mao, Tieliang Gong, Chen Li, and Erik Cambria.
\newblock Meta-based self-training and re-weighting for aspect-based sentiment analysis.
\newblock {\em IEEE Transactions on Affective Computing}, 14(3):1731--1742, 2022.

\bibitem{blending}
Erik Cambria, Newton Howard, Jane Hsu, and Amir Hussain.
\newblock Sentic blending: Scalable multimodal fusion for continuous interpretation of semantics and sentics.
\newblock In {\em {IEEE SSCI}}, pages 108--117, 2013.

\bibitem{somarathna2023exploring}
Rukshani Somarathna, Don~Samitha Elvitigala, Yijun Yan, Aaron~J Quigley, and Gelareh Mohammadi.
\newblock Exploring user engagement in immersive virtual reality games through multimodal body movements.
\newblock In {\em Proceedings of the 29th ACM Symposium on Virtual Reality Software and Technology}, pages 1--8, 2023.

\bibitem{yu2024artificial}
Joanne Yu, Astrid Dickinger, Kevin Kam~Fung So, and Roman Egger.
\newblock Artificial intelligence-generated virtual influencer: Examining the effects of emotional display on user engagement.
\newblock {\em Journal of Retailing and Consumer Services}, 76:103560, 2024.

\bibitem{wu2024promise}
Jialun Wu, Xinyao Yu, Kai He, Zeyu Gao, and Tieliang Gong.
\newblock Promise: A pre-trained knowledge-infused multimodal representation learning framework for medication recommendation.
\newblock {\em Information Processing \& Management}, 61(4):103758, 2024.

\bibitem{he2021construction}
Kai He, Lixia Yao, JiaWei Zhang, Yufei Li, and Chen Li.
\newblock Construction of genealogical knowledge graphs from obituaries: Multitask neural network extraction system.
\newblock {\em Journal of medical Internet research}, 23(8):e25670, 2021.

\bibitem{malandri2023convxai}
Lorenzo Malandri, Fabio Mercorio, Mario Mezzanzanica, and Navid Nobani.
\newblock Convxai: a system for multimodal interaction with any black-box explainer.
\newblock {\em Cognitive Computation}, 15(2):613--644, 2023.

\bibitem{liukno}
Hao Liu, Runguo Wei, Geng Tu, Jiali Lin, Dazhi Jiang, and Erik Cambria.
\newblock Knowing what and why: Causal emotion entailment for emotion recognition in conversations.
\newblock {\em Expert Systems with Applications}, 2025.

\bibitem{lialea}
Jiazhen Liang, Wai Li, Qingshan Zhong, Jun Huang, Dazhi Jiang, and Erik Cambria.
\newblock Learning chain for clause awareness: Triplex-contrastive learning for emotion recognition in conversations.
\newblock {\em Neural Computing and Applications}, 2025.

\bibitem{zambrano2023opportunistic}
Juan~M Zambrano~Chaves, Andrew~L Wentland, Arjun~D Desai, Imon Banerjee, Gurkiran Kaur, Ramon Correa, Robert~D Boutin, David~J Maron, Fatima Rodriguez, Alexander~T Sandhu, et~al.
\newblock Opportunistic assessment of ischemic heart disease risk using abdominopelvic computed tomography and medical record data: a multimodal explainable artificial intelligence approach.
\newblock {\em Scientific reports}, 13(1):21034, 2023.

\bibitem{he2025survey}
Kai He, Rui Mao, Qika Lin, Yucheng Ruan, Xiang Lan, Mengling Feng, and Erik Cambria.
\newblock A survey of large language models for healthcare: from data, technology, and applications to accountability and ethics.
\newblock {\em Information Fusion}, page 102963, 2025.

\bibitem{wu2023megacare}
Jialun Wu, Kai He, Rui Mao, Chen Li, and Erik Cambria.
\newblock {MEGACare: K}nowledge-guided multi-view hypergraph predictive framework for healthcare.
\newblock {\em Information Fusion}, 100:101939, 2023.

\bibitem{kim2023aobert}
Kyeonghun Kim and Sanghyun Park.
\newblock Aobert: All-modalities-in-one bert for multimodal sentiment analysis.
\newblock {\em Information Fusion}, 92:37--45, 2023.

\bibitem{maauec}
Yuansheng Ma, Dong Zhang, Shoushan Li, Erik Cambria, and Guodong Zhou.
\newblock {UECO}: Unified editing chain for efficient appearance transfer with multimodality-guided diffusion.
\newblock {\em Expert Systems with Applications}, 270:126510, 2025.

\bibitem{wancim}
Rui Wang, Chaopeng Guo, Erik Cambria, Imad Rida, Haochen Yuan, Md~Jalil Piran, Yichen Feng, Xianxun Zhu, and Mairie de~Compiegne.
\newblock {CIME}: Contextual interaction-based multimodal emotion analysis with enhanced semantic information.
\newblock {\em The Journal of Supercomputing}, 2025.

\bibitem{zadeh2017tensor}
Amir Zadeh, Minghai Chen, Soujanya Poria, Erik Cambria, and Louis-Philippe Morency.
\newblock Tensor fusion network for multimodal sentiment analysis.
\newblock In {\em Proceedings of the 2017 Conference on Empirical Methods in Natural Language Processing}, pages 1103--1114, 2017.

\bibitem{lin2024has}
Qika Lin, Yifan Zhu, Xin Mei, Ling Huang, Jingying Ma, Kai He, Zhen Peng, Erik Cambria, and Mengling Feng.
\newblock Has multimodal learning delivered universal intelligence in healthcare? a comprehensive survey.
\newblock {\em Information Fusion}, page 102795, 2024.

\bibitem{castro2019towards}
Santiago Castro, Devamanyu Hazarika, Ver{\'o}nica P{\'e}rez-Rosas, Roger Zimmermann, Rada Mihalcea, and Soujanya Poria.
\newblock Towards multimodal sarcasm detection (an \_obviously\_ perfect paper).
\newblock In {\em Proceedings of the 57th Annual Meeting of the Association for Computational Linguistics}, pages 4619--4629, 2019.

\bibitem{xu2018attngan}
Tao Xu, Pengchuan Zhang, Qiuyuan Huang, Han Zhang, Zhe Gan, Xiaolei Huang, and Xiaodong He.
\newblock Attngan: Fine-grained text to image generation with attentional generative adversarial networks.
\newblock In {\em Proceedings of the IEEE conference on computer vision and pattern recognition}, pages 1316--1324, 2018.

\bibitem{nguyen2019multi}
Duy-Kien Nguyen and Takayuki Okatani.
\newblock Multi-task learning of hierarchical vision-language representation.
\newblock In {\em Proceedings of the IEEE/CVF Conference on Computer Vision and Pattern Recognition}, pages 10492--10501, 2019.

\bibitem{han2021improving}
Wei Han, Hui Chen, and Soujanya Poria.
\newblock Improving multimodal fusion with hierarchical mutual information maximization for multimodal sentiment analysis.
\newblock In {\em Proceedings of the 2021 Conference on Empirical Methods in Natural Language Processing}, pages 9180--9192, 2021.

\bibitem{zhang2023learning}
Haoyu Zhang, Yu~Wang, Guanghao Yin, Kejun Liu, Yuanyuan Liu, and Tianshu Yu.
\newblock Learning language-guided adaptive hyper-modality representation for multimodal sentiment analysis.
\newblock In {\em Proceedings of the 2023 Conference on Empirical Methods in Natural Language Processing}, pages 756--767, 2023.

\bibitem{dong2023robust}
Yuxin Dong, Tieliang Gong, Shujian Yu, Hong Chen, and Chen Li.
\newblock Robust and fast measure of information via low-rank representation.
\newblock In {\em Proceedings of the AAAI Conference on Artificial Intelligence}, volume~37, pages 7450--7458, 2023.

\bibitem{yu2019multivariate}
Shujian Yu, Luis Gonzalo~Sanchez Giraldo, Robert Jenssen, and Jose~C Principe.
\newblock Multivariate~extension~of~matrix-based~r{\'e}nyi's~$\alpha$-order~entropy~functional.
\newblock {\em IEEE transactions on pattern analysis and machine intelligence}, 42(11):2960--2966, 2019.

\bibitem{xu2022different}
Bo~Xu, Shizhou Huang, Ming Du, Hongya Wang, Hui Song, Chaofeng Sha, and Yanghua Xiao.
\newblock Different data, different modalities! reinforced data splitting for effective multimodal information extraction from social media posts.
\newblock In {\em Proceedings of the 29th International Conference on Computational Linguistics}, pages 1855--1864, 2022.

\bibitem{sun2023efficient}
Licai Sun, Zheng Lian, Bin Liu, and Jianhua Tao.
\newblock Efficient multimodal transformer with dual-level feature restoration for robust multimodal sentiment analysis.
\newblock {\em IEEE Transactions on Affective Computing}, 2023.

\bibitem{nagrani2021attention}
Arsha Nagrani, Shan Yang, Anurag Arnab, Aren Jansen, Cordelia Schmid, and Chen Sun.
\newblock Attention bottlenecks for multimodal fusion.
\newblock {\em Advances in neural information processing systems}, 34:14200--14213, 2021.

\bibitem{pandey2023progress}
Ananya Pandey and Dinesh~Kumar Vishwakarma.
\newblock Progress, achievements, and challenges in multimodal sentiment analysis using deep learning: A survey.
\newblock {\em Applied Soft Computing}, page 111206, 2023.

\bibitem{tsai2019multimodal}
Yao-Hung~Hubert Tsai, Shaojie Bai, Paul~Pu Liang, J~Zico Kolter, Louis-Philippe Morency, and Ruslan Salakhutdinov.
\newblock Multimodal transformer for unaligned multimodal language sequences.
\newblock In {\em Proceedings of the conference. Association for computational linguistics. Meeting}, volume 2019, page 6558. NIH Public Access, 2019.

\bibitem{yu2022hierarchical}
Jianfei Yu, Kai Chen, and Rui Xia.
\newblock Hierarchical interactive multimodal transformer for aspect-based multimodal sentiment analysis.
\newblock {\em IEEE Transactions on Affective Computing}, 2022.

\bibitem{park2016multimodal}
Sunghyun Park, Han~Suk Shim, Moitreya Chatterjee, Kenji Sagae, and Louis-Philippe Morency.
\newblock Multimodal analysis and prediction of persuasiveness in online social multimedia.
\newblock {\em ACM Transactions on Interactive Intelligent Systems (TiiS)}, 6(3):1--25, 2016.

\bibitem{cai2015convolutional}
Guoyong Cai and Binbin Xia.
\newblock Convolutional neural networks for multimedia sentiment analysis.
\newblock In {\em Natural Language Processing and Chinese Computing: 4th CCF Conference, NLPCC 2015, Nanchang, China, October 9-13, 2015, Proceedings 4}, pages 159--167. Springer, 2015.

\bibitem{poria2016fusing}
Soujanya Poria, Erik Cambria, Newton Howard, Guang-Bin Huang, and Amir Hussain.
\newblock Fusing audio, visual and textual clues for sentiment analysis from multimodal content.
\newblock {\em Neurocomputing}, 174:50--59, 2016.

\bibitem{cheng2021multimodal}
Junyan Cheng, Iordanis Fostiropoulos, Barry Boehm, and Mohammad Soleymani.
\newblock Multimodal phased transformer for sentiment analysis.
\newblock In {\em Proceedings of the 2021 Conference on Empirical Methods in Natural Language Processing}, pages 2447--2458, 2021.

\bibitem{hoang2022multimodal}
Tuan Hoang, Thanh-Toan Do, Tam~V Nguyen, and Ngai-Man Cheung.
\newblock Multimodal mutual information maximization: A novel approach for unsupervised deep cross-modal hashing.
\newblock {\em IEEE Transactions on Neural Networks and Learning Systems}, 34(9):6289--6302, 2022.

\bibitem{hu2024survey}
Shizhe Hu, Zhengzheng Lou, Xiaoqiang Yan, and Yangdong Ye.
\newblock A survey on information bottleneck.
\newblock {\em IEEE Transactions on Pattern Analysis and Machine Intelligence}, 2024.

\bibitem{tishby2015deep}
Naftali Tishby and Noga Zaslavsky.
\newblock Deep learning and the information bottleneck principle.
\newblock In {\em 2015 ieee information theory workshop (itw)}, pages 1--5. IEEE, 2015.

\bibitem{shwartz2017opening}
Ravid Shwartz-Ziv and Naftali Tishby.
\newblock Opening the black box of deep neural networks via information.
\newblock {\em arXiv preprint arXiv:1703.00810}, 2017.

\bibitem{alemi2016deep}
Alexander~A Alemi, Ian Fischer, Joshua~V Dillon, and Kevin Murphy.
\newblock Deep variational information bottleneck.
\newblock In {\em International Conference on Learning Representations}, 2016.

\bibitem{achille2018information}
Alessandro Achille and Stefano Soatto.
\newblock Information dropout: Learning optimal representations through noisy computation.
\newblock {\em IEEE transactions on pattern analysis and machine intelligence}, 40(12):2897--2905, 2018.

\bibitem{higgins2017beta}
Irina Higgins, Loic Matthey, Arka Pal, Christopher~P Burgess, Xavier Glorot, Matthew~M Botvinick, Shakir Mohamed, and Alexander Lerchner.
\newblock beta-vae: Learning basic visual concepts with a constrained variational framework.
\newblock {\em ICLR (Poster)}, 3, 2017.

\bibitem{amjad2019learning}
Rana~Ali Amjad and Bernhard~C Geiger.
\newblock Learning representations for neural network-based classification using the information bottleneck principle.
\newblock {\em IEEE transactions on pattern analysis and machine intelligence}, 42(9):2225--2239, 2019.

\bibitem{wan2021multi}
Zhibin Wan, Changqing Zhang, Pengfei Zhu, and Qinghua Hu.
\newblock Multi-view information-bottleneck representation learning.
\newblock In {\em Proceedings of the AAAI conference on artificial intelligence}, volume~35, pages 10085--10092, 2021.

\bibitem{mai2022multimodal}
Sijie Mai, Ying Zeng, and Haifeng Hu.
\newblock Multimodal information bottleneck: Learning minimal sufficient unimodal and multimodal representations.
\newblock {\em IEEE Transactions on Multimedia}, 2022.

\bibitem{renyi1961measures}
Alfr{\'e}d R{\'e}nyi.
\newblock On measures of entropy and information.
\newblock In {\em Proceedings of the fourth Berkeley symposium on mathematical statistics and probability, volume 1: contributions to the theory of statistics}, volume~4, pages 547--562. University of California Press, 1961.

\bibitem{sanchez2014measures}
Luis Sanchez~Giraldo, Murali Rao, and Jose Principe.
\newblock Measures of entropy from data using infinitely divisible kernels.
\newblock {\em Information Theory, IEEE Transactions on}, 61, 11 2012.

\bibitem{tishby2000information}
Naftali Tishby, Fernando~C Pereira, and William Bialek.
\newblock The information bottleneck method.
\newblock {\em arXiv preprint physics/0004057}, 2000.

\bibitem{peng2018variational}
Xue~Bin Peng, Angjoo Kanazawa, Sam Toyer, Pieter Abbeel, and Sergey Levine.
\newblock Variational discriminator bottleneck: Improving imitation learning, inverse rl, and gans by constraining information flow.
\newblock {\em arXiv preprint arXiv:1810.00821}, 2018.

\bibitem{goyal2019infobot}
Anirudh Goyal, Riashat Islam, Daniel Strouse, Zafarali Ahmed, Matthew Botvinick, Hugo Larochelle, Yoshua Bengio, and Sergey Levine.
\newblock Infobot: Transfer and exploration via the information bottleneck.
\newblock {\em arXiv preprint arXiv:1901.10902}, 2019.

\bibitem{thomas2006elements}
MTCAJ Thomas and A~Thomas Joy.
\newblock {\em Elements of information theory}.
\newblock Wiley-Interscience, 2006.

\bibitem{lin2023ps}
Han Lin, Pinglu Zhang, Jiading Ling, Zhenguo Yang, Lap~Kei Lee, and Wenyin Liu.
\newblock Ps-mixer: A polar-vector and strength-vector mixer model for multimodal sentiment analysis.
\newblock {\em Information Processing \& Management}, 60(2):103229, 2023.

\bibitem{devlin2018bert}
Jacob Devlin, Ming-Wei Chang, Kenton Lee, and Kristina Toutanova.
\newblock Bert: Pre-training of deep bidirectional transformers for language understanding.
\newblock {\em arXiv preprint arXiv:1810.04805}, 2018.

\bibitem{imotions2017facet}
iMotions.
\newblock Facial expression analysis.
\newblock \url{https://www.imotions.com/}, 2017.

\bibitem{openface2018}
Tadas Baltrusaitis, Amir Zadeh, Yao~Chong Lim, and Louis-Philippe Morency.
\newblock Openface 2.0: Facial behavior analysis toolkit.
\newblock In {\em 2018 13th IEEE International Conference on Automatic Face \& Gesture Recognition (FG 2018)}, pages 59--66, 2018.

\bibitem{degottex2014covarep}
Gilles Degottex, John Kane, Thomas Drugman, Tuomo Raitio, and Stefan Scherer.
\newblock Covarep---a collaborative voice analysis repository for speech technologies.
\newblock In {\em 2014 ieee international conference on acoustics, speech and signal processing (icassp)}, pages 960--964. IEEE, 2014.

\bibitem{mcfee2015librosa}
Brian McFee, Colin Raffel, Dawen Liang, Daniel~PW Ellis, Matt McVicar, Eric Battenberg, and Oriol Nieto.
\newblock librosa: Audio and music signal analysis in python.
\newblock In {\em SciPy}, pages 18--24, 2015.

\bibitem{bhatia2006infinitely}
Rajendra Bhatia.
\newblock Infinitely divisible matrices.
\newblock {\em The American Mathematical Monthly}, 113(3):221--235, 2006.

\bibitem{lanczos1950iteration}
Cornelius Lanczos.
\newblock An iteration method for the solution of the eigenvalue problem of linear differential and integral operators.
\newblock 1950.

\bibitem{hazarika2020misa}
Devamanyu Hazarika, Roger Zimmermann, and Soujanya Poria.
\newblock Misa: Modality-invariant and-specific representations for multimodal sentiment analysis.
\newblock In {\em Proceedings of the 28th ACM international conference on multimedia}, pages 1122--1131, 2020.

\bibitem{kingma2014auto}
Diederik~P Kingma and Max Welling.
\newblock Auto-encoding variational bayes.
\newblock {\em stat}, 1050:1, 2014.

\bibitem{li2017collaborative}
Xiaopeng Li and James She.
\newblock Collaborative variational autoencoder for recommender systems.
\newblock In {\em Proceedings of the 23rd ACM SIGKDD international conference on knowledge discovery and data mining}, pages 305--314, 2017.

\bibitem{blei2017variational}
David~M Blei, Alp Kucukelbir, and Jon~D McAuliffe.
\newblock Variational inference: A review for statisticians.
\newblock {\em Journal of the American statistical Association}, 112(518):859--877, 2017.

\bibitem{yang2022multimodal}
Bo~Yang, Lijun Wu, Jinhua Zhu, Bo~Shao, Xiaola Lin, and Tie-Yan Liu.
\newblock Multimodal sentiment analysis with two-phase multi-task learning.
\newblock {\em IEEE/ACM Transactions on Audio, Speech, and Language Processing}, 30:2015--2024, 2022.

\bibitem{zadeh2016mosi}
Amir Zadeh, Rowan Zellers, Eli Pincus, and Louis-Philippe Morency.
\newblock Mosi: multimodal corpus of sentiment intensity and subjectivity analysis in online opinion videos.
\newblock {\em arXiv preprint arXiv:1606.06259}, 2016.

\bibitem{zadeh2018multimodal}
AmirAli~Bagher Zadeh, Paul~Pu Liang, Soujanya Poria, Erik Cambria, and Louis-Philippe Morency.
\newblock Multimodal language analysis in the wild: Cmu-mosei dataset and interpretable dynamic fusion graph.
\newblock In {\em Proceedings of the 56th Annual Meeting of the Association for Computational Linguistics (Volume 1: Long Papers)}, pages 2236--2246, 2018.

\bibitem{yu2020ch}
Wenmeng Yu, Hua Xu, Fanyang Meng, Yilin Zhu, Yixiao Ma, Jiele Wu, Jiyun Zou, and Kaicheng Yang.
\newblock Ch-sims: A chinese multimodal sentiment analysis dataset with fine-grained annotation of modality.
\newblock In {\em Proceedings of the 58th annual meeting of the association for computational linguistics}, pages 3718--3727, 2020.

\bibitem{niu2016sentiment}
Teng Niu, Shiai Zhu, Lei Pang, and Abdulmotaleb El~Saddik.
\newblock Sentiment analysis on multi-view social data.
\newblock In {\em MultiMedia Modeling: 22nd International Conference, MMM 2016, Miami, FL, USA, January 4-6, 2016, Proceedings, Part II 22}, pages 15--27. Springer, 2016.

\bibitem{wu2021graph}
Jianfeng Wu, Sijie Mai, and Haifeng Hu.
\newblock Graph capsule aggregation for unaligned multimodal sequences.
\newblock In {\em Proceedings of the 2021 international conference on multimodal interaction}, pages 521--529, 2021.

\bibitem{yuan2021transformer}
Ziqi Yuan, Wei Li, Hua Xu, and Wenmeng Yu.
\newblock Transformer-based feature reconstruction network for robust multimodal sentiment analysis.
\newblock In {\em Proceedings of the 29th ACM International Conference on Multimedia}, pages 4400--4407, 2021.

\bibitem{yu2021learning}
Wenmeng Yu, Hua Xu, Ziqi Yuan, and Jiele Wu.
\newblock Learning modality-specific representations with self-supervised multi-task learning for multimodal sentiment analysis.
\newblock In {\em Proceedings of the AAAI conference on artificial intelligence}, volume~35, pages 10790--10797, 2021.

\bibitem{mai2022hybrid}
Sijie Mai, Ying Zeng, Shuangjia Zheng, and Haifeng Hu.
\newblock Hybrid contrastive learning of tri-modal representation for multimodal sentiment analysis.
\newblock {\em IEEE Transactions on Affective Computing}, 14(3):2276--2289, 2022.

\bibitem{hwang2023self}
Yewon Hwang and Jong-Hwan Kim.
\newblock Self-supervised unimodal label generation strategy using recalibrated modality representations for multimodal sentiment analysis.
\newblock In {\em Findings of the Association for Computational Linguistics: EACL 2023}, pages 35--46, 2023.

\bibitem{huang2024tmbl}
Jiehui Huang, Jun Zhou, Zhenchao Tang, Jiaying Lin, and Calvin Yu-Chian Chen.
\newblock Tmbl: Transformer-based multimodal binding learning model for multimodal sentiment analysis.
\newblock {\em Knowledge-Based Systems}, 285:111346, 2024.

\bibitem{zadeh2018memory}
Amir Zadeh, Paul~Pu Liang, Navonil Mazumder, Soujanya Poria, Erik Cambria, and Louis-Philippe Morency.
\newblock Memory fusion network for multi-view sequential learning.
\newblock In {\em Proceedings of the AAAI conference on artificial intelligence}, volume~32, 2018.

\bibitem{mai2024meta}
Sijie Mai, Yu~Zhao, Ying Zeng, Jianhua Yao, and Haifeng Hu.
\newblock Meta-learn unimodal signals with weak supervision for multimodal sentiment analysis.
\newblock {\em arXiv preprint arXiv:2408.16029}, 2024.

\bibitem{xu2017analyzing}
Nan Xu.
\newblock Analyzing multimodal public sentiment based on hierarchical semantic attentional network.
\newblock In {\em 2017 IEEE international conference on intelligence and security informatics (ISI)}, pages 152--154. IEEE, 2017.

\bibitem{xu2018co}
Nan Xu, Wenji Mao, and Guandan Chen.
\newblock A co-memory network for multimodal sentiment analysis.
\newblock In {\em The 41st international ACM SIGIR conference on research \& development in information retrieval}, pages 929--932, 2018.

\bibitem{yang2020image}
Xiaocui Yang, Shi Feng, Daling Wang, and Yifei Zhang.
\newblock Image-text multimodal emotion classification via multi-view attentional network.
\newblock {\em IEEE Transactions on Multimedia}, 23:4014--4026, 2020.

\bibitem{yang2021multimodal}
Xiaocui Yang, Shi Feng, Yifei Zhang, and Daling Wang.
\newblock Multimodal sentiment detection based on multi-channel graph neural networks.
\newblock In {\em Proceedings of the 59th Annual Meeting of the Association for Computational Linguistics and the 11th International Joint Conference on Natural Language Processing (Volume 1: Long Papers)}, pages 328--339, 2021.

\bibitem{li2022clmlf}
Zhen Li, Bing Xu, Conghui Zhu, and Tiejun Zhao.
\newblock Clmlf: A contrastive learning and multi-layer fusion method for multimodal sentiment detection.
\newblock In {\em Findings of the Association for Computational Linguistics: NAACL 2022}, pages 2282--2294, 2022.

\bibitem{chen2024holistic}
Junyu Chen, Jie An, Hanjia Lyu, Christopher Kanan, and Jiebo Luo.
\newblock Holistic visual-textual sentiment analysis with prior models.
\newblock In {\em 2024 IEEE 7th International Conference on Multimedia Information Processing and Retrieval (MIPR)}, pages 196--202. IEEE, 2024.

\bibitem{wang2024multimodal}
Hongbin Wang, Chun Ren, and Zhengtao Yu.
\newblock Multimodal sentiment analysis based on cross-instance graph neural networks.
\newblock {\em Applied Intelligence}, 54(4):3403--3416, 2024.

\bibitem{giraldo2014measures}
Luis Gonzalo~Sanchez Giraldo, Murali Rao, and Jose~C Principe.
\newblock Measures of entropy from data using infinitely divisible kernels.
\newblock {\em IEEE Transactions on Information Theory}, 61(1):535--548, 2014.

\bibitem{cohen1988}
Jacob Cohen.
\newblock {\em Statistical Power Analysis for the Behavioral Sciences}.
\newblock Routledge, 1988.

\bibitem{wei2019eda}
Jason Wei and Kai Zou.
\newblock Eda: Easy data augmentation techniques for boosting performance on text classification tasks.
\newblock In {\em Proceedings of the 2019 Conference on Empirical Methods in Natural Language Processing and the 9th International Joint Conference on Natural Language Processing (EMNLP-IJCNLP)}, pages 6382--6388, 2019.

\bibitem{liu2018efficient}
Zhun Liu, Ying Shen, Varun~Bharadhwaj Lakshminarasimhan, Paul~Pu Liang, Amir Zadeh, and Louis-Philippe Morency.
\newblock Efficient low-rank multimodal fusion with modality-specific factors.
\newblock {\em arXiv preprint arXiv:1806.00064}, 2018.

\bibitem{mai2020modality}
Sijie Mai, Haifeng Hu, and Songlong Xing.
\newblock Modality to modality translation: An adversarial representation learning and graph fusion network for multimodal fusion.
\newblock In {\em {AAAI}}, pages 164--172, 2020.

\bibitem{su2020msaf}
Lang Su, Chuqing Hu, Guofa Li, and Dongpu Cao.
\newblock Msaf: Multimodal split attention fusion.
\newblock {\em arXiv preprint arXiv:2012.07175}, 2020.

\bibitem{ghosal2018contextual}
Deepanway Ghosal, Md~Shad Akhtar, Dushyant Chauhan, Soujanya Poria, Asif Ekbal, and Pushpak Bhattacharyya.
\newblock Contextual inter-modal attention for multi-modal sentiment analysis.
\newblock In {\em proceedings of the 2018 conference on empirical methods in natural language processing}, pages 3454--3466, 2018.

\bibitem{van2008visualizing}
Laurens Van~der Maaten and Geoffrey Hinton.
\newblock Visualizing data using t-sne.
\newblock {\em Journal of machine learning research}, 9(11), 2008.

\end{thebibliography}



\end{document}